\documentclass[nohyperref]{article}

\usepackage{microtype}
\usepackage{graphicx}
\usepackage{subfigure}
\usepackage{booktabs} % for professional tables

\usepackage{hyperref}

\usepackage[accepted]{icml2022}

\usepackage{amsmath}
\usepackage{amssymb}
\usepackage{mathtools}
\usepackage{amsthm}

\usepackage[capitalize,noabbrev]{cleveref}

\theoremstyle{plain}

\theoremstyle{definition}

\theoremstyle{remark}

\usepackage[disable,textsize=tiny]{todonotes}
\usepackage{color,soul}
\usepackage{multirow}
\usepackage[T1]{fontenc}

\usepackage{bm}

\usepackage{cancel}

\icmltitlerunning{\nickname: CAD Construction Sequence Generation with Disentangled Codebooks}

\newcommand{\nickname}{SkexGen}

\begin{document}

\twocolumn[
\icmltitle{\nickname: Autoregressive Generation of CAD Construction \\Sequences with Disentangled Codebooks}



\icmlsetsymbol{equal}{*}

\begin{icmlauthorlist}
\icmlauthor{Xiang Xu}{sfu}
\icmlauthor{Karl D.D. Willis}{ad}
\icmlauthor{Joseph G. Lambourne}{ad}\\
\icmlauthor{Chin-Yi Cheng}{ad}
\icmlauthor{Pradeep Kumar Jayaraman}{ad}
\icmlauthor{Yasutaka Furukawa}{sfu}
\end{icmlauthorlist}

\icmlaffiliation{sfu}{Simon Fraser University, Canada}
\icmlaffiliation{ad}{Autodesk Research}

\icmlcorrespondingauthor{Xiang Xu}{xuxiangx@sfu.ca}

\icmlkeywords{Generative Model, CAD, VQ-VAE, Transformer, Autoregressive}

\vskip 0.3in
]



\printAffiliationsAndNotice{}  % leave blank if no need to mention equal contribution

% Joe: Adding some macros for adding comments for our internal use
\newif\ifshowcomments
% Toggle this flag to show/hide comments
%\showcommentstrue
\showcommentsfalse
%\ifshowcomments
%\newcommand{\joe}[1]{{\color{red}{[Joe: #1]}}}
%\newcommand{\karl}[1]{{\color{purple}{[Karl: #1]}}}
%\newcommand{\pradeep}[1]{{\color{cyan}{[Pradeep: #1]}}}
%\newcommand{\sam}[1]{{\color{red}{#1}}}
%\newcommand{\yasu}[1]{{\color{blue}{[Yasu: #1]}}}
%\newcommand{\chinyi}[1]{{\color{blue}{[Chin-Yi: #1]}}}
%\else
%\newcommand{\joe}[1]{}
%\newcommand{\karl}[1]{}
%\newcommand{\pradeep}[1]{}
%\newcommand{\sam}[1]{}
%\newcommand{\yasu}[1]{}
%\newcommand{\chinyi}[1]{}
%\fi

\newcommand{\mysubsubsection}[1]{\vspace{0.1cm} \noindent {\bf #1}:}
\newcommand{\mysubsubsectionA}[1]{\vspace{0.1cm} \noindent {\bf #1}}
\newcommand{\mysubsubsubsection}[1]{\vspace{0.1cm} \noindent {\bf #1}:}

\newcommand{\etal}{\textit{et al.}}

%%%%%%%%% ABSTRACT
\begin{abstract}
    We present \nickname, a novel autoregressive generative model for computer-aided design (CAD) construction sequences containing sketch-and-extrude modeling operations. Our model utilizes distinct Transformer architectures to encode topological, geometric, and extrusion variations of construction sequences into disentangled codebooks. Autoregressive Transformer decoders generate CAD construction sequences sharing certain properties specified by the codebook vectors. Extensive experiments demonstrate that our disentangled codebook representation generates diverse and high-quality CAD models, enhances user control, and enables efficient exploration of the design space. The code is available at \url{https://samxuxiang.github.io/skexgen}.
\end{abstract}

%%%%%%%%% BODY TEXT
\section{Introduction}
\label{sec:intro}

Professional designers generate a diverse set of computer aided design (CAD) models with topological or geometric variations while achieving a design goal. In early concept design, designers explore shapes with aesthetic and functional advantages. In mechanical design, designers optimize the physical properties of a part to maximize strength while minimizing weight and manufacturing cost. 

Training a computational agent with the capabilities of a professional designer is an extremely challenging machine learning task. Designers traditionally rely on parametric CAD models that allow one to correct small mistakes~\cite{camba2016parametric}, optimize geometric properties~\cite{Jurgen2004}, and generate a product family by altering just a few parameters~\cite{chakrabarti2011computer,maher1996formalising}. However, a parametric CAD model is brittle and fails over large design changes such as topological modifications. Furthermore, the construction of a parametric CAD model itself requires the expertise of a professional designer. 

With the advance of deep learning, neural networks empower a family of exciting new methodologies~\cite{wu2021deepcad,willis2021engineering,Ganin2021ComputerAidedDA} towards an intelligent system capable of diverse generation while understanding a design goal and allowing user control. This paper pushes the frontier of the state-of-the-art by introducing ``{\nickname}'', a \textbf{sk}etch-and-\textbf{ex}trude \textbf{gen}erative model for CAD construction sequences that enhances variations in design generation while enabling effective control and exploration of the design space.

{\nickname} is a novel autoregressive generative model that uses discrete codebooks \cite{oord2018neural} for CAD model generation. We employ a sketch-and-extrude
modeling language to describe CAD construction sequences, where a sketch operation creates 2D primitives and an extrude operation lifts and combines them into 3D. Transformer encoders learn a disentangled latent representation as three codebooks, capturing topological, geometric, and extrusion variations. Given codebook vectors, autoregressive Transformer decoders generate sketch-and-extrude construction sequences, which are processed into a CAD model.

We evaluate {\nickname} on a large-scale sketch-and-extrude dataset~\cite{wu2021deepcad}. Qualitative and quantitative evaluations against multiple baselines and state-of-the-art methods demonstrate that our method generates more realistic and diverse CAD models, while allowing effective control and efficient exploration of the design space not possible with prior approaches. We make the following contributions.

\noindent $\bullet$ {\nickname} architecture that autoregressively generates high quality and diverse CAD construction sequences.

\noindent $\bullet$ Disentangled codebooks, which encode topological, geometric, and extrusion variations of construction sequences, enabling effective control and exploration of designs.

\noindent $\bullet$ Extensive qualitative and quantitative evaluations on public benchmarks, demonstrating state-of-the-art performance.

\section{Related Work}
\label{sec:rw}

\mysubsubsection{Constructive Solid Geometry}
3D shapes can be expressed as a constructive solid geometry (CSG) tree, where parametric primitives, such as cuboids, spheres, and cones, are combined together with Boolean operations.
%, such as union and subtraction. 
This lightweight representation has been used extensively for reconstruction tasks in combination with program synthesis~\cite{du2018inversecsg,nandi2017programming,nandi2018functional}, neurally-guided program synthesis~\cite{sharma2018csgnet,ellis2019write,tian2019learning}, unsupervised learning~\cite{kania2020ucsg}, and with specialized parametric primitives~\cite{chen2020bsp,yu2022capri}. Although CSG is a convenient representation, parametric CAD remains the most prevalent paradigm for mechanical design and makes extensive use of the sketch-and-extrude modeling operations instead.

\mysubsubsection{Construction Sequence Generation}
A precursor to sketch-and-extrude construction sequence generation is PolyGen~\cite{nash2020polygen}, where $n$-gon mesh vertices and faces are predicted using Transformers~\cite{vaswani2017attention} and pointer networks~\cite{vinyals2015pointer}.
Large-scale datasets of parametric CAD construction sequences~\cite{Ari2020,willis2020fusion} have opened the door to learning directly from the modeling operations by CAD users. \cite{willis2020fusion} and \cite{xu2021zone} predict the sequence of extrude operations for partial recovery of the construction sequence without the underlying sketch information. Predicting sequences of sketch primitives (e.g., line, arc, circle) is a critical building block that forms the 2D basis of CAD and can be readily extended to 3D with the addition of the extrude operation.

The Transformer architecture has enabled sketch/sketch-and-extrude construction sequence generation in recent~\cite{willis2021engineering,wu2021deepcad} and concurrent~\cite{para2021sketchgen,seff2021vitruvion,Ganin2021ComputerAidedDA} work. 
Although these approaches can produce a diverse range of shapes, providing user control over an existing design is more elusive. To be useful for real world CAD applications, the designer needs a way to influence the generated shape.  
One approach is to condition the network on user-provided images~\cite{Ganin2021ComputerAidedDA}, point clouds~\cite{wu2021deepcad}, or hand-drawn sketches~\cite{seff2021vitruvion}.
However, this approach simply converts an existing design into a CAD construction sequence representation. Instead we present a novel approach for exploring the space of related designs with separate control over topology and geometry.

\mysubsubsection{Codebook Architectures}
Codebooks have proven effective on a number of image and audio generation tasks since their introduction by \cite{oord2018neural}, improving the diversity of generated images \cite{razavi2019generating} and providing additional user control~\cite{esser2021taming}. They are particularly suited to encoding CAD modeling sequences due to their high structural regularity.

\section{Sketch-and-Extrude Construction Sequence}
\label{sec:def}

\begin{figure*}[ht]
\begin{center}
\centerline{\includegraphics[width=\linewidth]{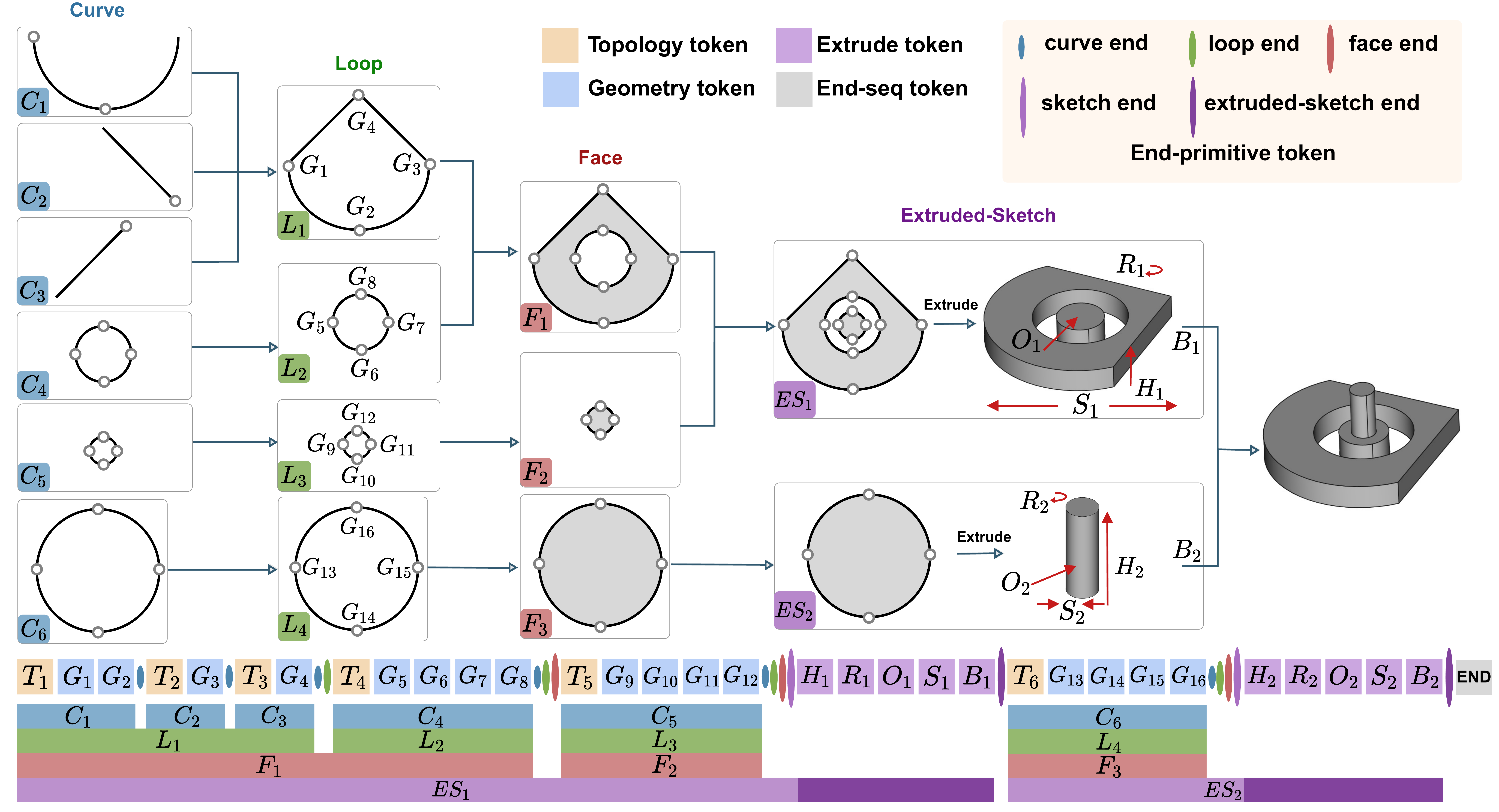}}
\caption{A sample sketch-and-extrude model is illustrated as a primitive hierarchy and a code sequence. It consists of two sketches, which are formed by faces, loops, and curves. The sequence starts by a topology token ($T_1$), indicating the start of a curve (type arc). Two geometry tokens follow ($G_1$ and $G_2$), each of which stores a 2D point coordinate. Note that a circle is defined by four points with four geometry tokens (e.g., $G_5$--$G_8$). An arc is defined by three points (e.g., $G_1$--$G_3$) but with two tokens, where the third point is specified by the next curve (or the first curve when a loop is closed). Similarly, a line is defined by start point (e.g., $G_4$). $G_8$ is the last point for curve ($C_4$), loop ($L_2$), and face ($F_1$). Therefore, the geometry token ($G_8$) is followed by end-primitive tokens for curve, loop, and face.}
\label{fig:data_format}
\end{center}
\vskip -0.3in
\end{figure*}

We define a sketch-and-extrude construction sequence representation as a hierarchy of primitives, building on TurtleGen~\cite{willis2021engineering} and DeepCAD~\cite{wu2021deepcad} with several modifications to make the representation more expressive and amenable for learning (See Figure~\ref{fig:data_format}).

\mysubsubsection{Primitive Hierarchy}
A ``curve'' (i.e., line, arc, or circle) is the lowest-level primitive. A ``loop'' is a closed path, consisting of one (i.e., circle) or multiple curves (e.g., line-arc-line).
A ``face'' is a 2D area bounded by loops, which is new in our representation. Precisely, a face is defined by one outer loop and some number of inner loops as holes, a convention in many CAD systems \cite{lee2001partialentity}.
A ``sketch'' is formed by one or multiple faces. An ``extruded-sketch'' is a 3D volume, formed by extruding a sketch. A ``sketch-and-extrude'' model is formed by multiple extruded-sketches via Boolean operations (i.e., intersection, union, and subtraction). Note that the DeepCAD~\cite{wu2021deepcad} representation does not have a face primitive and cannot represent a sketch with multiple faces (e.g., $ES_1$ in Figure~\ref{fig:data_format}).

\mysubsubsection{Construction Sequence}
Following DeepCAD~\cite{wu2021deepcad}, we represent a sketch-and-extrude CAD model using a sequence with five types of tokens: 1) A topology token indicates a curve type (line/arc/circle); 2) A geometry token contains a point coordinate; 3) An end-primitive token indicates the end of a primitive (curve/loop/face/sketch/extruded-sketch); 4) An extrusion token contains parameters associated with the extrusion and Boolean operations; and 5) An end-sequence token indicates the end of the sequence. Figure~\ref{fig:data_format} illustrates a sample primitive hierarchy and a sequence. We provide the full language specification in Appendix~\ref{sec:appendix_sequence}. 
\section{SkexGen Architecture}
\label{sec:method}
\begin{figure*}[ht]
\begin{center}
\centerline{\includegraphics[width=\linewidth]{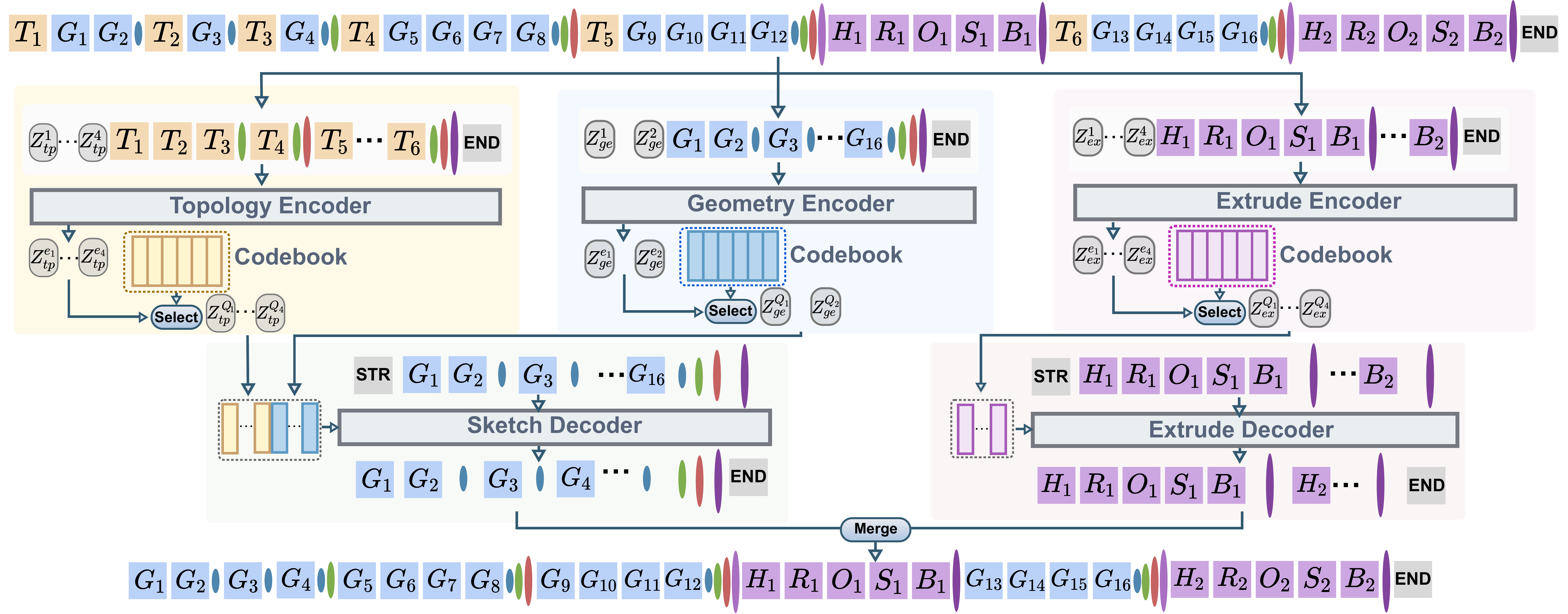}}
\caption{The SkexGen architecture has two branches. The sketch branch (left) has two encoders that learn topological and geometric variations of sketches in two codebooks. An autoregressive decoder generates the sketch subsequence given codebook vectors. The extrude branch (right) has an encoder and a decoder that learns variations of extrude and Boolean operations. The two branches are trained independently. Another autoregressive decoder learns to select effective combinations of codes from the three codebooks (not shown).}
\label{fig:model}
\end{center}
\vskip -0.3in
\end{figure*}

{\nickname} is an autoregressive generative model that learns variations of sketch-and-extrude models with three disentangled codebooks in two network branches. Figure~\ref{fig:model} visualizes the {\nickname} architecture. The ``Sketch'' branch learns topological and geometric variations of 2D sketches, while the ``Extrude'' branch learns variations of 3D extrusions (e.g., directions). Both branches are similar and this section explains the sketch branch with two encoders and one decoder. Details of the extrude branch are in Appendix~\ref{sec:appendix_extrude_branch}. 

\subsection{Topology Encoder} \label{sec:topology_encoder}
The topology encoder takes a subsequence of the input, where the token is either 1) a topology token ($T$), which indicates one of the $3$ curve types (line/arc/circle); 2) an end-primitive token ($E$) for one of the $3$ primitive types (loop/face/sketch); or 3) An end-sequence token ($End$), indicating the end of the sequence. 
Accordingly, a token is initialized with a one-hot vector of dimension $7(=3+3+1)$.

\mysubsubsection{Embeddings}
The one-hot vector is transformed into a $d_\text{E}=256$ dimensional embedding. We consider a topology token $T$ in the subsequence where $h^{tp}_T$ is the 7-dim one hot vector and $i_T$ denotes its index in the input subsequence. Its embedding vector is computed as:
\begin{eqnarray}
\label{eq:embedding}
T \leftarrow \mathbf{W^{tp}} h^{tp}_T + \mathbf{p}^{(i_T)},
\end{eqnarray}
where $\mathbf{W^{tp}}\in \mathbb{R}^{d_\text{E} \times 7}$ denotes a learnable matrix. $\mathbf{p}^{(i_T)} \in \mathbb{R}^{d_\text{E}}$ denotes the learnable position encoding at index $i_T$ of the topology subsequence\footnote{With abuse of notation, $T$ denotes a sequence token and its latent embedding. A bold-font indicates learnable parameters.}.

\mysubsubsection{Architecture}
The encoder is based on a standard Transformer architecture~\cite{vaswani2017attention} with four layer blocks, each of which contains a self-attention layer with eight heads, layer-normalization and feed-forward layers. Following Vision Transformer~\cite{Dosovitskiy2021AnII}, the input topology information is encoded into a ``code-token'', which is prepended to the input and initialized with a learnable embedding $Z_{tp}$. Let $Z^{e}_{tp}$ be the embedding of a code-token at the output of the encoder. The embedding $Z^{e}_{tp}$ is quantized to the closest code in the codebook of size $N$: $\{\mathbf{b}_{tp}^{(i)} ~ | ~ i=1, 2\cdots N\}$. The final code-token $Z^{Q}_{tp}$ after encoding and quantization is then passed to the decoder:
\begin{eqnarray}
Z^{Q}_{tp} \leftarrow \mathbf{b}_{tp}^{(k)},   \hspace{0.1cm} \text{where} \hspace{0.1cm} k = \mbox{argmin}_j \|Z^{e}_{tp} - \mathbf{b}_{tp}^{(j)}\|^2. 
\end{eqnarray}
Here we assumed one code-token for simplicity. In practice, the topology encoder has four code-tokens and makes four output codes $\Big(Z^{Q_{(1)}}_{tp},Z^{Q_{(2)}}_{tp},Z^{Q_{(3)}}_{tp},Z^{Q_{(4)}}_{tp}\Big)$. We tried different codebook sizes and found that $N=500$ achieves good results.

\subsection{Geometry Encoder} \label{sec:geometry_encoder}
The geometry encoder takes a subsequence of the input, where a token is either 1) a geometry token ($G$); 2) an end-primitive token ($E$) for one of the 4 primitive types (curve/face/loop/sketch); or 3) an end-sequence token ($End$).
The geometry token $G$ specifies a 2D point coordinate along a curve. Since a coordinate is numerical, we discretize sketches into $64\times 64$ (6 bits) pixels and consider $64^2$ possible pixel locations\footnote{6 bits yield enough precision for most CAD models as described in Vitruvion~\cite{seff2021vitruvion}. More bits bring little improvement at the cost of significantly more network parameters.}. Therefore, a one-hot vector of dimension $4101(=64^2+4+1)$ uniquely determines the token information. 

\mysubsubsection{Embeddings}
We follow Eq.~\ref{eq:embedding} and use a learnable matrix $\mathbf{W^{ge}}\in \mathbb{R}^{d_{\text{E}}\times 4101}$ together with position encoding to initialize input token embeddings. Token $E$ and $End$ are initialized similarly to the topology token $T$, by multiplying their one-hot vector with $\mathbf{W^{ge}}$ and adding the position encoding. Geometry tokens $G$ are initialized differently as:
\begin{eqnarray}
G &\leftarrow& \mathbf{W^{ge}} h^{ge}_G + 
\mathbf{W^{x}} h^{x}_G + 
\mathbf{W^{y}} h^{y}_G + \mathbf{p}^{(i_G)}.
\end{eqnarray}
$h^{ge}_G \in \mathbb{R}^{4101}$ denotes the one-hot vector.
The geometry tokens $G$ have additional coordinate embeddings where
$h^x_G, h^y_G \in \mathbb{R}^{64}$ is a one-hot vector indicating the x, y coordinate of the pixel. $\mathbf{W^{x}}, \mathbf{W^{y}} \in \mathbb{R}^{d_{\text{E}}\times 64}$ are learnable coordinate matrices. Coordinate embeddings are optional but further improve results in our experiments.

\mysubsubsection{Architecture}
Similar to the topology encoder, code-tokens in the geometry encoder produces the embedding $\{Z_{ge}^{e_{(i)}}\}$  and the quantized code $\{Z^{Q_{(i)}}_{ge}\}$. We use two code-tokens for the geometry encoder ($i=1,2$). Codebook size is $1000$.

\subsection{Sketch Decoder} \label{sec:decoder}
The sketch decoder takes as input the topology and geometry codebooks and  generates geometry tokens $G$ and end-primitive tokens $E$ (for curve/loop/face/sketch) to recover the sketch subsequence. Note that topology tokens $T$ are not generated, as they can be inferred from the number of geometry tokens within each curve (i.e., line/arc/circle have 1/2/4 $G$ tokens). This means that geometry encoder and sketch decoder have similar subsequence (see Figure~\ref{fig:model}).

\mysubsubsection{Input}
Given the past $k-1$ tokens, the autoregressive decoder predicts the conditional probability of the $k^\text{th}$ token. The training input sequences are shifted one to the right, with the ``start'' symbol (initialized by the position encoding) added at front. Since the types of possible tokens in the decoder are equivalent to those of the geometry encoder, we use the same one-hot encoding scheme of dimension $4101$ and also a learnable matrix of size $d_{\text{E}}\times 4101$ with position encoding to initialize the embedding vectors.

\mysubsubsection{Output} The decoder produces a subsequence ``shifted one to the left'', that is, predicting the original $k$ tokens in the input (See Figure~\ref{fig:model}). Let $K$ be a token in the output of sketch decoder, which has a $d_\text{E}$ dimensional embedding. We use a learnable matrix $\mathbf{W^{out}} \in \mathbb{R}^{4101\times d_\text{E}}$ to predict the 
%$4101$ dimensional 
probability likelihood over $4101$ classes as: 
%whose format is the same as the one-hot vector of the geometry encoder:
\begin{eqnarray}
h^{\text{out}}_{K} \leftarrow \text{softmax} \left( \mathbf{W^{out}} K \right).
\end{eqnarray}

\mysubsubsection{Cross-attention} 
The Transformer architecture takes four and two quantized codebook vectors from the topology and the geometry codebooks via cross-attention. To distinguish two different codebooks, we borrow the idea of position encoding and add learnable embedding vectors $\mathbf{p}^{(q_{tp})} \in \mathbb{R}^{4\times d_\text{E}} $ and $\mathbf{p}^{(q_{ge})} \in \mathbb{R}^{2\times d_\text{E}}$ to the topology code $\{Z^{Q_{(i)}}_{tp}\}$ and geometry code $\{Z^{Q_{(i)}}_{ge}\}$ respectively:
\begin{eqnarray}
Z^{Q_{(i)}}_{tp} + \mathbf{p}^{(q_{tp})} \quad \text{or} \quad
Z^{Q_{(i)}}_{ge} + \mathbf{p}^{(q_{ge})}. \label{eq:cross_attention}
\end{eqnarray}
The base network setting is the same as the encoder (i.e., 4 layer-blocks with 8 heads), except that it is autoregressive with masking (only previous tokens are attended).

\subsection{Training}
The topology encoder, geometry encoder, and sketch decoder are jointly trained with three loss functions:
\begin{eqnarray}
\label{eq:loss_fn}
&&\sum_K \text{CrossEntropy}(h^{\text{out}}_{K}, h^{\text{gt}}_{K}) + \\
&&\left| \left| sg\left(Z_{tp}^{e}\right) - \mathbf{b}_{tp} \right| \right|^2_2 + \beta \left| \left| Z_{tp}^{e} - sg\left(\mathbf{b}_{tp}\right) \right| \right|^2_2 +  \nonumber \\
&&\left| \left| sg\left(Z_{ge}^{e}\right) - \mathbf{b}_{ge} \right| \right|^2_2 + \beta\left|  \left| Z_{ge}^{e} - sg\left(\mathbf{b}_{ge}\right) \right| \right|^2_2.  \nonumber 
\end{eqnarray} 
The first line computes the sequence reconstruction loss, where $h^{\text{out}}_{K}$ is the predicted probability likelihood from the sketch decoder and $h^{\text{gt}}_{K}$ is the ground-truth one-hot vector. The second and the third lines are standard codebook and commitment losses used with VQ-VAE~\cite{oord2018neural}\footnote{
We use the exponential moving average updates of a decay rate 0.99, which was effective in VQ-VAE-2~\cite{razavi2019generating}.}. $sg$ denotes the stop-gradient operation, which is the identity
function in the forward pass but blocks gradients in the backward pass. $\beta$ scales the commitment loss and is set to 0.25. This ensures the encoder output commits to one code vector.
%and prevent it from changing frequently between different codes.  
We omit explicitly writing out the multiple code-tokens  within each encoder for simplicity.

Given a ground-truth subsequence, we run the two encoders and autoregressively run the decoder until the same number of tokens are generated. Instead of feeding back the generated tokens, we do teacher forcing and pass the ground-truth to the input of the decoder, which effectively allows us to train the decoder for a single step each.

\subsection {Generation}
\label{sec:codebook_prior}

{\nickname} generates CAD models in two steps: 1) code generation from the three codebooks; and 2) sketch-and-extrude construction sequence generation given the codes.

\mysubsubsection{Code generation}
Unlike a VAE, our quantized codes do not follow a Normal distribution. After using the trained topology, geometry, and extrude encoders to obtain codes from each training sample, we train a Transformer decoder that generates codes, i.e. selecting code indexes from each of the three codebooks. Our framework also allows conditional code generation with a minor modification to the architecture. For example, a ``topology conditioned code selector'' selects the compatible geometry and extrude codes given topology codes. We provide full details in Appendix~\ref{sec:appendix_code_selection}.

\mysubsubsection{Sequence generation}
Given the codes, the sketch and the extrude decoders autoregressively generate construction subsequences separately by nucleus sampling~\cite{holtzman2019curious}.
The subsequences are merged to form a complete sketch-and-extrude sequence, which is then parsed to boundary representation~\cite{weiler1986topological} using CAD software.

\section{Experiments}\label{sec:exp}
In this section we perform experiments to understand: 1) the ability of \nickname~to generate high quality and diverse results, 2) the level of control that codebooks enable over the generation process, and 3) the performance of \nickname~for applications such as design exploration and interpolation.

\subsection{Experiment Setup} \label{sec:exp_setup}

\mysubsubsection{Dataset}
We use the DeepCAD dataset~\cite{wu2021deepcad} which contains  178,238 sequences and a data split of $90\%$ train, $5\%$ validation, and $5\%$ test. Duplicates are removed in a similar manner to \citet{willis2021engineering}. We separately remove duplicate sketch subsequences and duplicate extrude subsequences. Invalid sketch-and-extrude operations are also removed. The resulting training dataset contains 74,584 sketch subsequences and 86,417 extrude subsequences. For experiments with single sketches, we extract sketches from all steps of the CAD construction sequences. This leads to 114,985 training samples after duplicate removal. 

\mysubsubsection{Implementation details}
\nickname~is implemented in PyTorch \cite{paszke2019pytorch} and trained on a RTX A5000. For fair comparison, we follow DeepCAD \cite{wu2021deepcad} and use a Transformer with four-layer blocks where each block contains eight attention heads, pre-layer normalization, and feed-forward dimension of $512$. The input embedding dimension is $256$. Dropout rate is $0.1$. During training, we use the Adam optimizer~\cite{kingma2014adam} with a learning rate of $0.001$. Linear warm-up and gradient clipping are used as in DeepCAD. We skip code quantization in the initial 25 epochs and find this stabilizes the codebook training. For data augmentation, we add small random noise to the coordinates of the geometry tokens. {\nickname} is trained for a total of $300$ epochs with a batch size of $128$. Maximum length is $200$ for sketch subsequences and $100$ for extrude subsequences. At test time, we use nucleus sampling to autoregressively sample the code selector and the decoders.

\begin{figure}[t]
\begin{center}
\includegraphics[width=0.99\columnwidth]{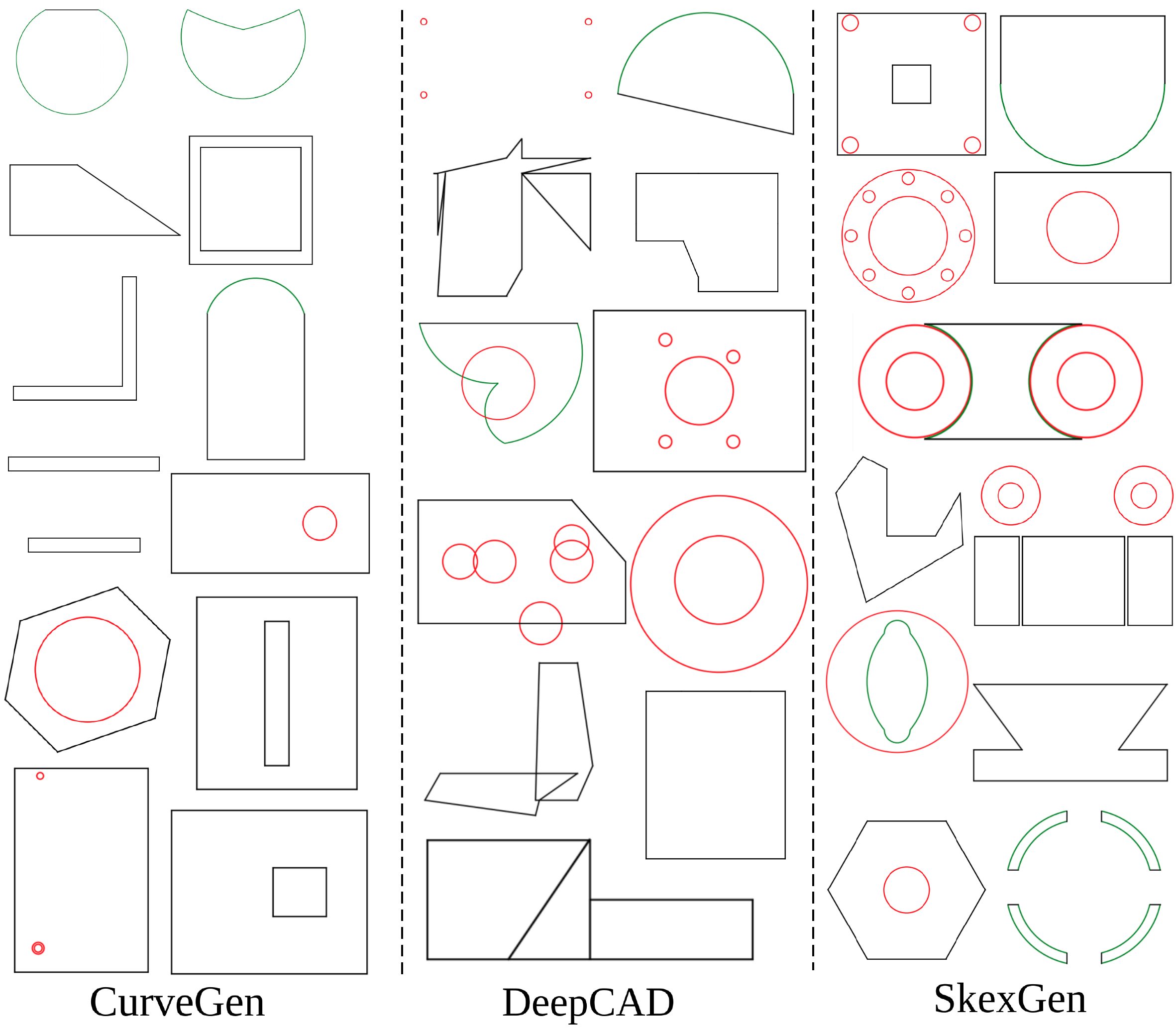}
\caption{
Random sketch generation results by CurveGen, DeepCAD, and {\nickname}. Circle is red, arc is green, line is black. Visually, SkexGen generates more realistic and complex shapes.
}
\label{fig:random_gen_sketch}
\end{center}
\vskip -0.2in
\end{figure}

\mysubsubsection{Metrics}
We use the following metrics for the quantitative evaluation. For 2D sketches, ``Fr\'{e}chet inception distance (FID)"~\cite{heusel2017gans} measures the generation fidelity. It compares mean and covariance of real and generated data distributions. Following~\cite{das2020beziersketch}, we use features from ResNet-18~\cite{he2016deep} pre-trained on a human sketch classification task~\cite{eitz2012hdhso}. For 3D CAD models, ``Coverage" (COV) is the percentage of real data that match generated data based on the closest Chamfer distance of 2,000 uniformly sampled points on the surface. 
``Minimum Matching Distance" (MMD) is the average minimum matching distance between a generated sample and its nearest neighbor in the real set. ``Jensen-Shannon Divergence" (JSD) is the similarity between the real and generated distributions based on the marginal point distribution. For both sketches and CAD models, the ``Novel'' score is the percentage of generated data that does not appear in the training set, the ``Unique'' score is the percentage of data that only appears once in the generated samples. We consider two data samples equal if all tokens in the sequence are the same after 6-bit quantization. Please refer to \cite{achlioptas2018learning,wu2021deepcad} for details.

%%%%%%%%%%%%%%%%%%%%%%%%%%%%%%%%%%%%%%%%%%%%%%%%%%%%%%%%%
\begin{table}[t]
\caption{Quantitative evaluations on the sketch generation task. We report the FID score to measure generation fidelity and the percentage of Unique and Novel.}
\label{tab:sketch_rand}
\begin{center}
\small
\begin{tabular}{@{}lccc@{}}
\toprule
                 & FID & Unique & Novel \\ 
                 & $\downarrow$  & \% $\uparrow$   & \% $\uparrow$ \\ \midrule                 
CurveGen         & 109.12          & 93.05            & 83.82           \\
DeepCAD          & 75.47           & 98.79            & 97.45           \\
%SkexGen w/o code & 29.31           & 95.20            & 81.62           \\
SkexGen          & 18.56           & 96.02            & 83.54           \\ \bottomrule
\end{tabular}
\end{center}
\vskip -0.2in
\end{table}

%%%%%%%%%%%%%%%%%%%%%%%%%%%%%%%%%%%%%%%%%%%%%%%%%%%%%%%%%

%%%%%%%%%%%%%%%%%%%% random cad %%%%%%%%%%%%%%%%%%%%%%%%%%
\begin{figure*}[t]
\includegraphics[width=\textwidth]{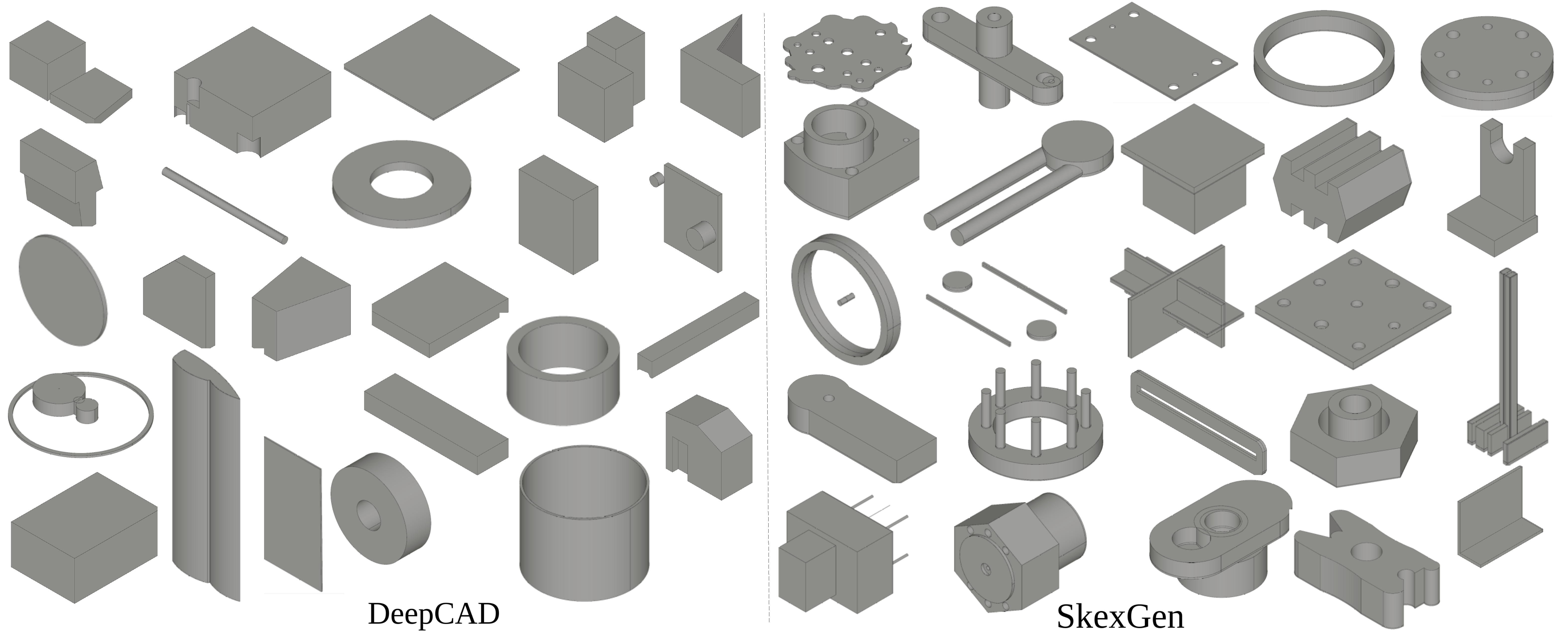}
\caption{Random sketch-and-extrude CAD model generation results by DeepCAD and {\nickname}. Visually, SkexGen generates more realistic and complex models.}
\label{fig:rand_gen_cad}
\vskip -0.1in
\end{figure*}

%%%%%%%%%%%%%%%%%%%%%%%%%%%%%%%%%%%%%%%%%%%%%%%%%%%%%%%%%

\subsection{Random Generation}
To assess the ability of \nickname~to generate high quality and diverse results, we randomly generate 20,000 samples each and compare \nickname~with four other baselines: CurveGen~\cite{willis2021engineering}, DeepCAD~\cite{wu2021deepcad}, SkexGen with a single codebook, and SkexGen with VAE. Other sketch generative models from concurrent work~\cite{para2021sketchgen,seff2021vitruvion,Ganin2021ComputerAidedDA} rely on sketch constraint labels, and ideally a sketch constraint solver, which makes them not directly comparable.

\mysubsubsection{Sketch generation} We report quantitative results for sketch generation in Table~\ref{tab:sketch_rand}. {\nickname} has by far the best FID score. The unique and novel percentage are similar or better than CurveGen, with the exception that novel percentage is lower than DeepCAD. Closer examination of the qualitative results in Figure~\ref{fig:random_gen_sketch} reveals that DeepCAD produces many invalid results containing self-intersecting curves and
non-watertight geometry. Since invalid results do not appear in the training data, they are all counted as novel and the score is high. The FID metric detects that the invalid data is far
from the ground truth distribution and consequently the DeepCAD FID score is much worse. The Unique metric for DeepCAD is only slightly better than SkexGen; we suspect this is due to the increased noise in the generated data.

 Overall, we find that sketches from {\nickname} are better in terms of quality with more complex shapes, fewer self-intersections, and stronger symmetry. CurveGen also generates good quality results, but with fewer complex arrangements of rectangles and circles. DeepCAD can produce more complex shapes than CurveGen but with a lot of noise. Additional visualizations of our generated sketches are available in Appendix~\ref{sec:appendix_experiment}.
 
%%%%%%%%%%%%%%%%%%%%%%%%%%%%%%%%%%%%%%%%%%%%%%%%%%%%%%%%%
\begin{table}[t]
\caption{Quantitative evaluations on the CAD generation task based on the Coverage (COV) percentage, Minimum Matching Distance (MMD),  Jensen-Shannon Divergence (JSD) and the percentage of Unique and Novel. 
The first row is the existing state-of-the-art DeepCAD~\cite{wu2021deepcad}. The last three rows provide an ablation study of SkexGen over two features: code disentanglement ($F_{d}$) and code quantization ($F_{q}$). 
The second from last row disables the code disentanglement (i.e., using a single codebook), resembling standard VQ-VAE. The third from last row further disables the code quantization, resembling standard VAE.
} 
\label{tab:cad_rand}
\begin{center}
\small
\setlength{\tabcolsep}{5pt}
\begin{tabular}{@{}lccccc@{}}
\toprule
         & COV   & MMD  & JSD  & Unique   & Novel    \\ 
         &  \% $\uparrow$  & $\downarrow$  & $\downarrow$  &\% $\uparrow$   & \% $\uparrow$    \\ \midrule
DeepCAD                 & 76.8 & 1.68 & 2.01 & 91.0 & 87.0 \\  \midrule
SkexGen - ($F_{d}, F_{q}$)   & 74.3 & 1.54 & 0.92 & 97.8 & 91.9  \\ 
SkexGen - ($F_d$)       & 80.4 & 1.55 & 1.12 & 99.8 & 99.3  \\
SkexGen                 & 83.6 & 1.48  & 0.81 & 99.9 & 99.8  \\
\bottomrule
\end{tabular}
%\small
%\begin{tabular}{@{}lccccc@{}}
%\toprule
%         & COV   & MMD  & JSD  & Unique   & Novel    \\ 
%         &  \% $\uparrow$  & $\downarrow$  & $\downarrow$  &\% $\uparrow$   & \% $\uparrow$    \\ \midrule
%DeepCAD                 & 76.75 & 1.68 & 2.01 & 91.04 & 86.98 \\  \midrule
%SkexGen - ($F_{d}, F_{q}$)   & 74.30 & 1.54 & 0.92 & 97.78 & 91.93  \\ 
%SkexGen - ($F_d$)       & 80.35 & 1.55 & 1.12 & 99.82 & 99.30  \\
%SkexGen                 & 83.60 & 1.48  & 0.81 & 99.86 & 99.87  \\
%%\begin{tabular}[c]{@{}l@{}}SkexGen - multi\\ - code \end{tabular}  & 74.30 & 1.54 & 0.92 & 97.78 & 91.93  \\ 
%\bottomrule
%\end{tabular}
\end{center}
\vskip -0.3in
\end{table}

%\begin{tabular}{@{}lcccccc@{}}
%\toprule
%         & COV   & MMD  & JSD  & Unq   & Nov   & Val   \\ \midrule
%VAE      & 74.30 & 1.54 & 0.92 & 97.78 & 91.93 & 85.54 \\
%VQVAE    & 80.35 & 1.55 & 1.12 & 99.82 & 99.30 & 77.80 \\
%DeepCAD  & 76.75 & 1.68 & 2.01 & 91.04 & 86.98 & 77.11 \\
%SkextGen & 83.60 & 1.48 & 0.81 & 100.0 & 99.9  & 75.34 \\ \bottomrule
%\end{tabular}
%%%%%%%%%%%%%%%%%%%%%%%%%%%%%%%%%%%%%%%%%%%%%%%%%%%%%%%%%

%%%%%%%%%%%%%%%%%%%%%%%%%%%%%%%%%%%%%%%%%%%%%%%%%%%%%%%%%

\begin{figure*}[t]
\centerline{\includegraphics[width=0.8\textwidth]{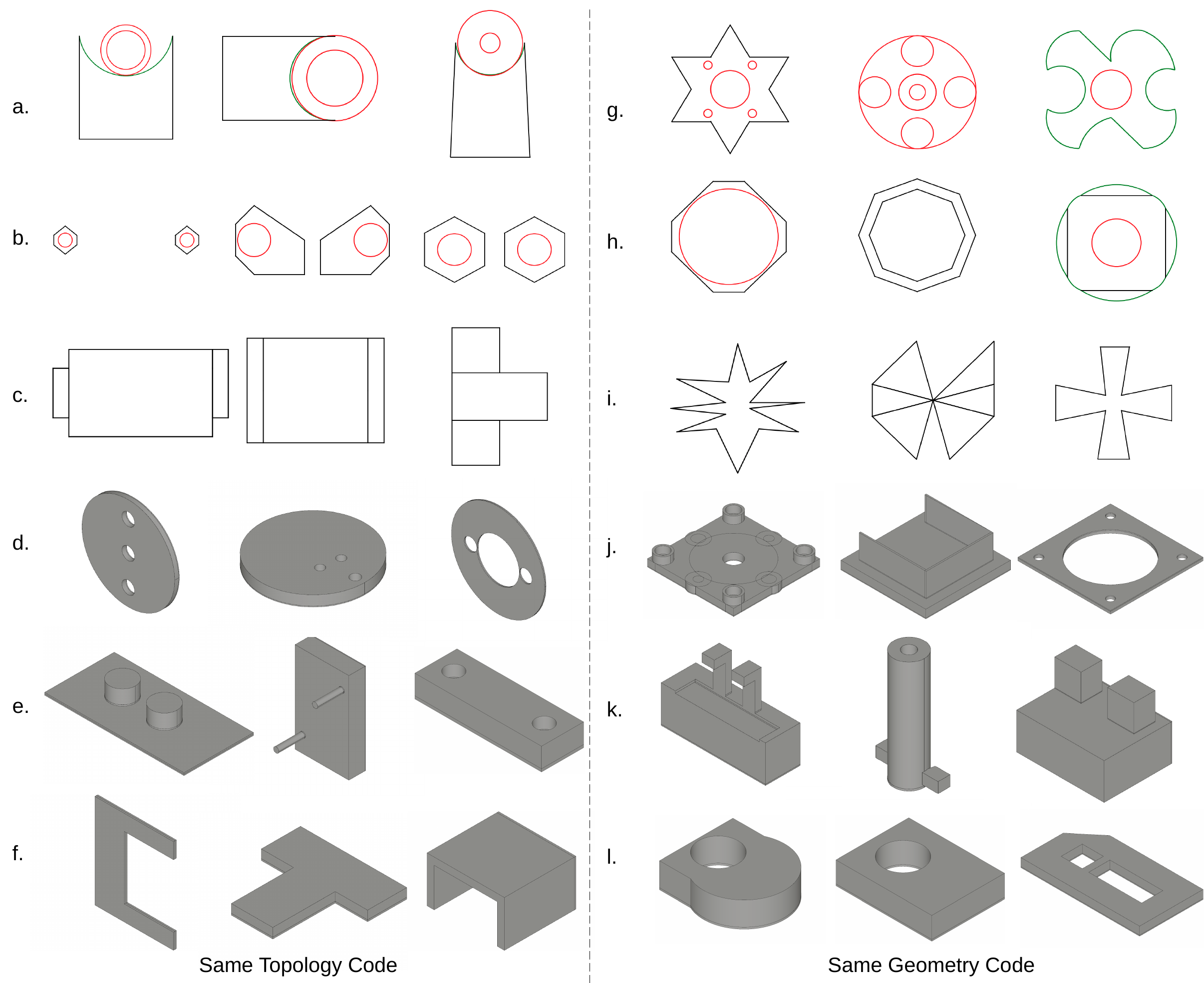}}
\caption{Controllable generation results. Each row shares the same topology (left) or the same geometry (right) codes. Other codes are randomly sampled by the code selector. Top rows are sketches and bottom rows are sketch-and-extrude CAD models. }
\label{fig:cond_code}
\vskip -0.1in
\end{figure*}

%%%%%%%%%%%%%%%%%%%%%%%%%%%%%%%%%%%%%%%%%%%%%%%%%%%%%%%%%

%%%%%%%%%%%%%%%%%%%%%%%%%%%%%%%%%%%%%%%%%%%%%%%%%%%%%%%%%

\begin{figure}[t]
\centerline{\includegraphics[width=0.99\columnwidth]{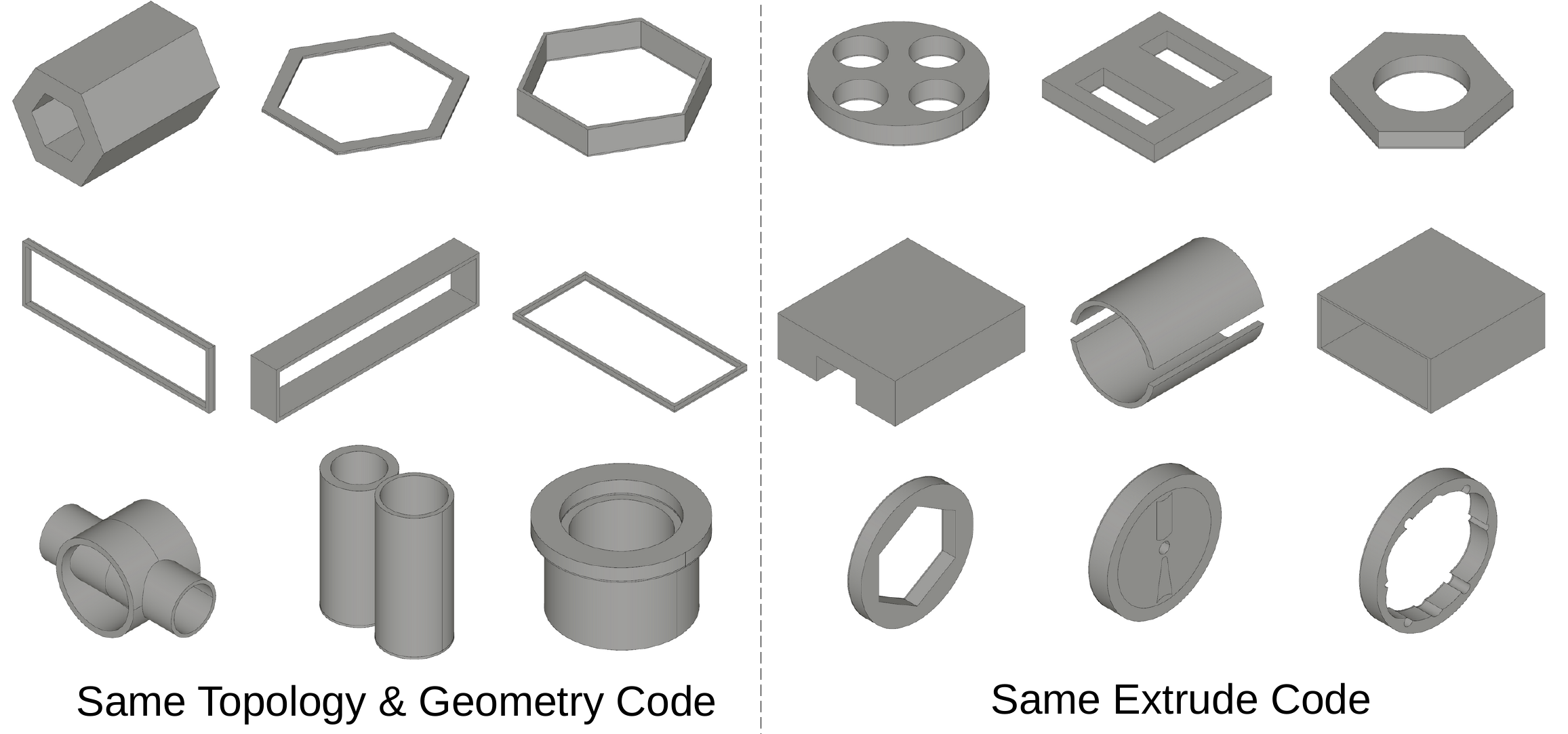}}
\caption{Controllable generation results. Each row shares the same topology and geometry (left) or the same extrude (right) codes. Other codes are randomly sampled by the code selector.
}
\label{fig:cond_extrude}
\vskip -0.2in
\end{figure}

%%%%%%%%%%%%%%%%%%%%%%%%%%%%%%%%%%%%%%%%%%%%%%%%%%%%%%%

\begin{figure*}
{\includegraphics[width=\textwidth]{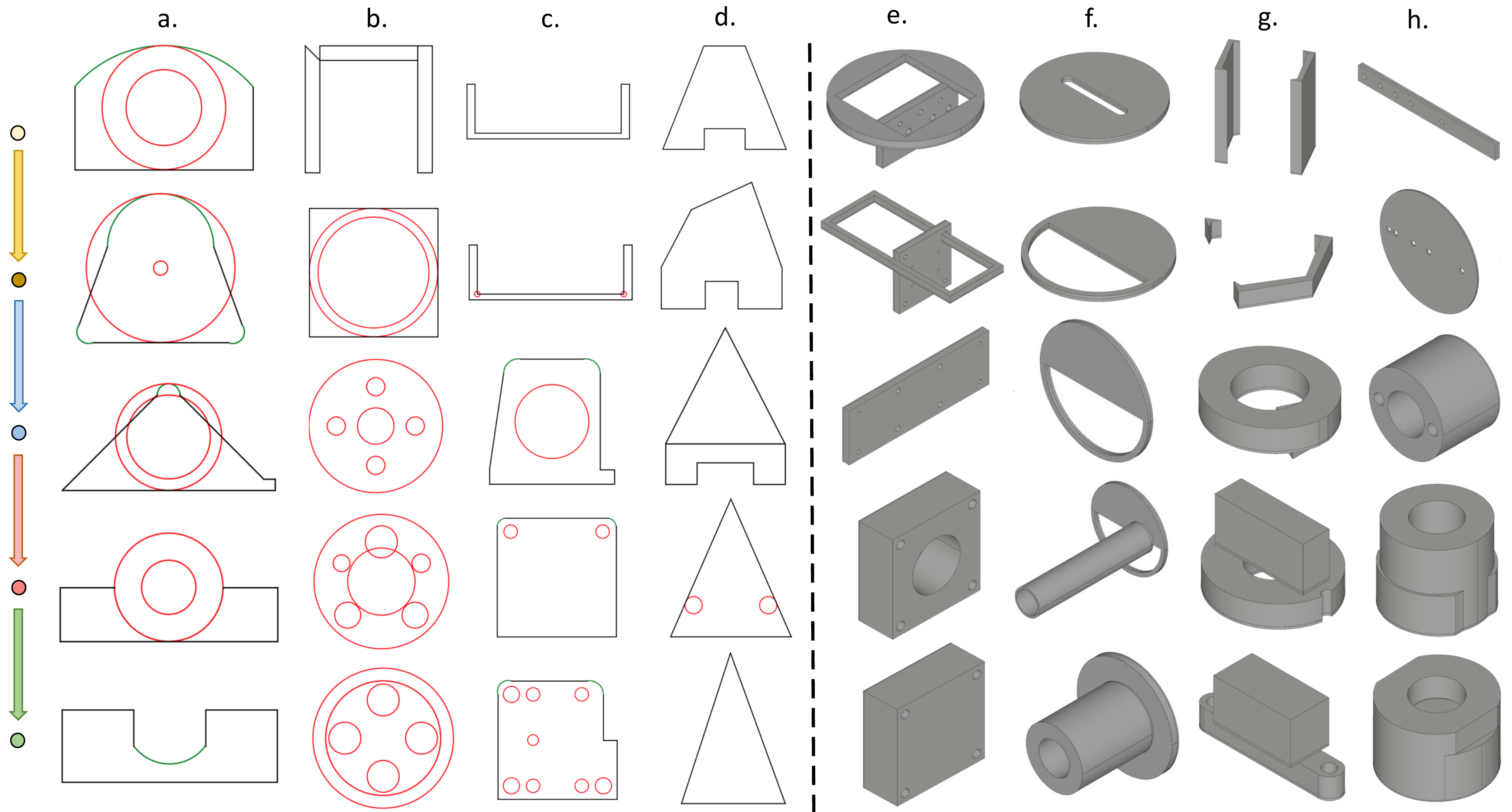}}
\vskip -0.1in
\caption{Design interpolation results for sketches (left) and CAD models (right). The top and the bottom rows show the source and target shapes, interpolated results appear in the middle rows.
}
\label{fig:inter_all}
\end{figure*}

\begin{figure*}
{\includegraphics[width=\textwidth]{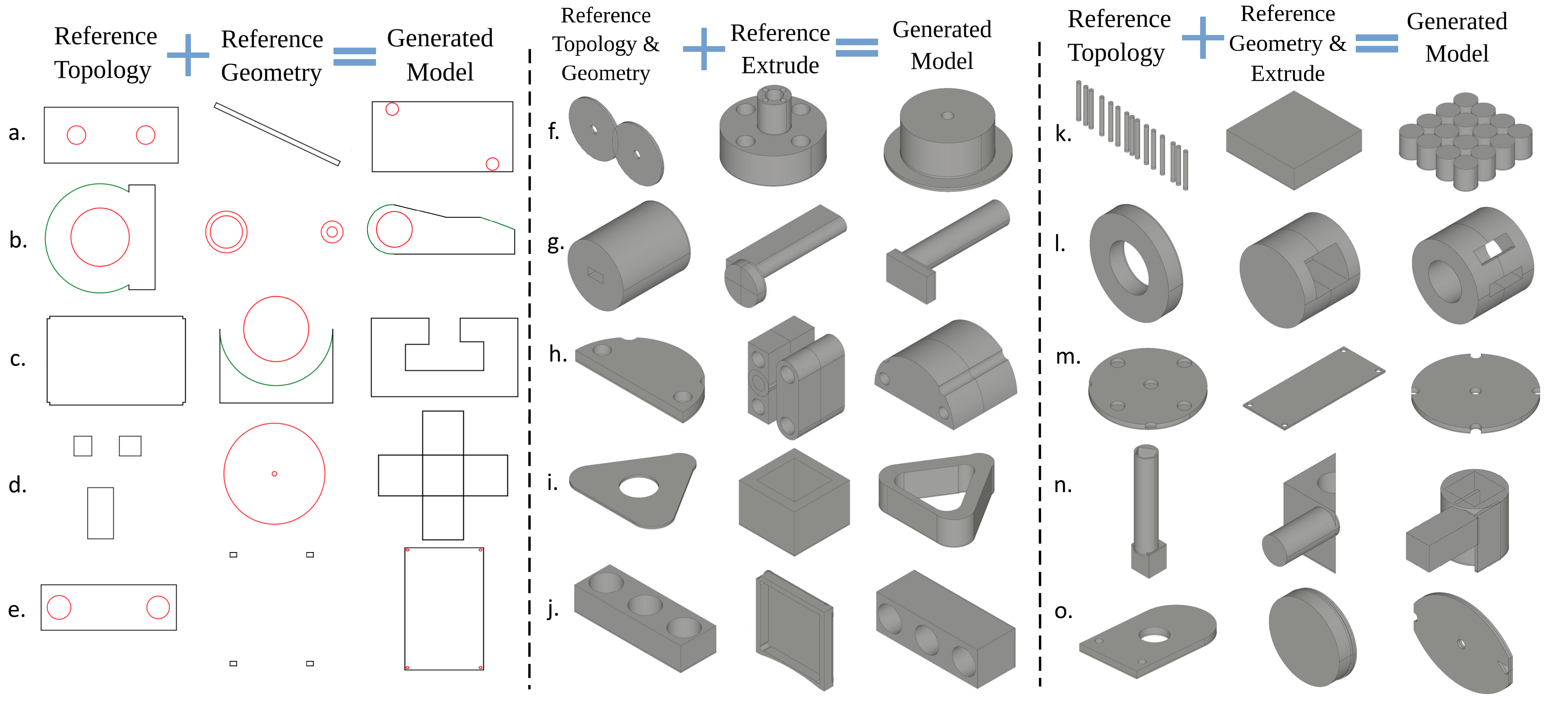}}
\vskip -0.1in
\caption{Topology, geometry, and extrude code mixing results. Given two reference CAD models or sketches (A and B), SkexGen can generate a new data whose topological property is similar to A and the geometric property is similar to B, by simply copying the topology codes from A, the geometry codes from B, and passing the mixed codes to the decoders for sampling. 
}
\label{fig:cond_all}
\end{figure*}

\mysubsubsection{CAD generation} 
Table~\ref{tab:cad_rand} provides quantitative evaluations on the sketch-and-extrude CAD model generation. We find that {\nickname} performs the best across all metrics. Qualitative results for {\nickname} and DeepCAD are shown in Figure~\ref{fig:rand_gen_cad} with additional results available in Appendix~\ref{sec:appendix_experiment}. {\nickname} generates CAD models that are considerably more complex, exhibit symmetries, and make frequent use of the \textit{arc} curve type, reminiscent of human design. Results from {\nickname} also contain frequent multi-step sketch-and-extrude sequences, whereas DeepCAD results are mostly single-step. The amount of similar shapes generated by DeepCAD is also high. The middle two rows in Table~\ref{tab:cad_rand} demonstrate the effectiveness of multiple disentangled codebooks. The generation quality decreases after reduction of multiple codebooks to just one and {\nickname} is similar to a VQ-VAE. Results are the worst when no codebook is used and {\nickname} effectively becomes a VAE.

\mysubsubsection{Running time} {\nickname} is slower than DeepCAD due to autoregressive sampling, taking 90 secs to generate 10,000 samples compared to 15 secs for DeepCAD. However, it is $2\times$ faster than CurveGen since CurveGen has two dependent autoregressive decoders.

\subsection{Controllable Generation}
Disentangled codebooks allow effective control and design exploration. The left of Figure~\ref{fig:cond_code} illustrates ``topology conditioned code selection'' results, where the conditioned topology codes are the same for each row and the other codes are obtained by nucleus sampling. For example, in row b, all sketches contain two separate faces, which consist of an inner circle and an outer loop of six lines. The geometric properties such as face size and distance between the two faces vary. The right of Figure~\ref{fig:cond_code} similarly shows ``geometry conditioned code selection'' results, where the geometry codes are fixed for each row. In row g, all the sketches have roughly the same size and arrangement of curves, while the topology of the outer loops are different, from lines (left) to circle (middle) to arcs (right). For CAD models in the bottom rows, extrude operations further vary, influencing the 3D layout of the models (e.g., row e). Figure~\ref{fig:cond_extrude} has more controllable generation results with different conditional code selectors, demonstrating the capability of SkexGen.

To quantitatively measure the disentanglement between the three codebooks, we follow the evaluation in $\beta$-VAE \cite{higgins2016beta}. A pair of sketch-and-extrude sequences are generated by the decoders by keeping one of the topology, geometry or extrude tokens the same, and sampling the rest. A small Transformer-based classifier is trained to identify which code is fixed using the average pairwise difference in the encoded latent space over all pairs of data. The classification accuracy for {\nickname} is $99.8 \pm 0.1\%$.

\subsection{Applications}
We demonstrate two applications enabled by the proposed {\nickname} system.

\mysubsubsectionA{Interpolation} allows intuitive design exploration. Given a pair of models, we follow the same encoding and decoding process in Sect.~\ref{sec:method}, except that we linearly interpolate their codes before quantization ($\{Z^{e}\}$) and produce one model.
%we interpolate their codes before quantization ($\{Z^{e}\}$) and generate a sequence of interpolated models. 
Precisely, we 1) use the encoders to extract 10 codes (4 topology $\{Z^{e}_{tp}\}$, 2 geometry $\{Z^{e}_{ge}\}$, and 4 extrude $\{Z^{e}_{ex}\}$) from each model; 2) linearly interpolate them; and 3) perform the code quantization and autoregressive sequence generation to produce an interpolated model.
%we use the encoders to extract 10 $Z^{e}$ codes before quantization (4 topology, 2 geometry, and 4 extrude) from each model.
%
%We interpolate each code and 
%of the 10 codes between the two models, redo the codebook selection process and pass the new $Z^{Q}$ to the decoder for sampling. 
%
Note that we take the token of the highest score during the autoregressive generation for consistent interpolation instead of the nucleus sampling.
%We with the top-1 sampling instead of the nucleus sampling for better visualization. 
Figure~\ref{fig:inter_all} shows interesting topological and geometric changes over the interpolation. In column b, the topology of the sketch changes from a set of 12 lines to a set of 4 lines and 2 circles, and finally to a set of 6 circles. The radius of the middle circle gradually increases until enclosing the four smaller ones.
In column h, a rectangular solid gradually morphs into a circular hollow disk. Note that the results are often not ``smooth'', but our interpolation tasks are particularly challenging, requiring many topological changes that are discrete in nature.

\mysubsubsectionA{Topology, geometry, and extrude code mixing} enables generation results not seen in previous work. In Figure~\ref{fig:cond_all} we mix the topology, geometry, or extrude codes from one data with those of another to produce a hybrid result. For example, in row a, the generated sketch has two inner circles as in the reference topology while the placement of the two circles resembles that of the reference geometry. In row k, the reference topology indicates many cylinders and the reference geometry indicates a square, resulting in many cylinders in the arrangement of a square.

\section{Conclusion}
\label{sec:conclude}
We introduced \nickname, a novel generative model for CAD construction sequences that enhances topological and geometric variations to enable better design control and exploration. \nickname~is a step towards an intelligent system capable of generating diverse CAD models while comprehending design goals and allowing user control.

\section*{Acknowledgements}
The research is supported by NSERC Discovery Grants, NSERC Discovery Grants Accelerator Supplements, and DND/NSERC Discovery Grants.


\clearpage
\bibliography{main}
\bibliographystyle{icml2022}

%%%%%%%%%%%%%%%%%%%%%%%%%%%%%%%%%%%%%%%%%%%%%%%%%%%%%%%%%%%%%%%%%%%%%%%%%%%%%%%
%%%%%%%%%%%%%%%%%%%%%%%%%%%%%%%%%%%%%%%%%%%%%%%%%%%%%%%%%%%%%%%%%%%%%%%%%%%%%%%
% APPENDIX
%%%%%%%%%%%%%%%%%%%%%%%%%%%%%%%%%%%%%%%%%%%%%%%%%%%%%%%%%%%%%%%%%%%%%%%%%%%%%%%
%%%%%%%%%%%%%%%%%%%%%%%%%%%%%%%%%%%%%%%%%%%%%%%%%%%%%%%%%%%%%%%%%%%%%%%%%%%%%%%
\newpage
\appendix
\onecolumn

\section{Construction Sequence: Full Specification}\label{sec:appendix_sequence}

A construction sequence must follow certain rules to be a valid sketch-and-extrude model. Instead of providing a full specification as a context free grammar, which is possible, we describe rules with a few bullet points.
\begin{itemize}
\item A sketch-and-extrude model consists of multiple extruded-sketches.
\item A sketch consists of multiple faces.
\item A face consists of multiple loops. 
\item A loop consists of multiple curves. 
\item A curve is either a line, an arc, or a circle, which consist of 2, 3, or 4 points, respectively.
\item A loop with a circle can not contain additional curves since it is already a closed path.
\item A point is represented by a geometry token.
\item An end-primitive token appears at the end of each primitive (curve, line, face, loop, sketch, or extruded-sketch).
\item When a face consists of multiple loops, the first loop defines the external boundary, and the remaining loops define internal loops (i.e., holes).
\end{itemize}

\section{Extrude Branch Architecture} \label{sec:appendix_extrude_branch}
The extrude branch takes a subsequence of the input, where a token is either 1) An extrude token, which is one of the five types ($H$,$R$,$O$, $S$, $B$); 2) An end-primitive token ($E$) for the extruded-sketch primitive type; and 3) An end-sequence token ($End$).

\begin{itemize}
    \item
    %$H$ indicates the offsets of the top and the bottom extruded-sketch plane from the reference plane. A negative values means it is below the reference plane, a positive value means it is above it.
$H$ indicates the displacements of the top and the bottom planes from the reference plane in which a sketch is extruded to form a solid.
%heights of the planes as displacements from the reference plane.
%specifies two values.
Since we know that the two values are necessary, we repeat $H$ tokens twice in the sequence instead of having one token encode two values. We quantize each height value into 64 bins and represent the value as a $64$ dimensional one-hot vector.

\item
$R$ is a 3D rotation of the extrusion direction. Looking through $3\times 3$ rotation matrices in our datasets, each entry of the rotation matrix has a value of either -1, 0, or 1. These three values cover almost all rotations from DeepCAD dataset (>99\%). Similarly to $H$, we repeat $R$ nine times in the sequence, where each token indicates one of the three values as a 3 dimensional one-hot vector. 
%Therefore, $R$ indicates one of the above three rotations as a 3 dimensional one-hot vector instead of storing a full rotation parameter set.

\item
$O$ is a 3D translation vector applied to the extruded solid. Similarly to $H$, we know that three values are necessary, and we repeat $O$ three times in the sequence. We quantize each height value into 64 bins and represent the value as a $64$ dimensional one-hot vector.

\item
$S$ indicates a sketch scaling factor consisting of three values: the center of scaling as a 2D coordinate and the uniform scaling factor. Similarly, we quantize each value into 64 bins as a $64$ dimensional one-hot vector.

\item
$B$ indicates one of the three Boolean operations (intersection, union, or subtraction) and is represented by a $3$ dimensional one-hot vector.
\end{itemize}

\mysubsubsection{Embeddings}
Figure~\ref{fig:data_format} in the main paper shows that an extrude subsequence consists of 7 types of tokens including $End$. In fact, an extrude subsequence after flatten consists of $19(=2+9+3+3+1+1)$ individual tokens: $[HHRRRRRRRRROOOSSSBE]$ (not counting $End$). Each token is represented by a $72(=64+3+3+1+1)$ dimensional one-hot vector, where numeric tokens ($H$, $O$, $S$) share the 64 bins.

\mysubsubsection{Extrude encoder} The encoder is the same as the topology or geometry encoder except that a one-hot vector is of 72 dimensional instead of 4101. Similar to Eq.~\ref{eq:cross_attention}, we also add an additional learnable embedding which distinguish the token types (e.g. the 9 rotation matrix tokens VS the 3 translation tokens).

\mysubsubsection{Extrude decoder} The decoder is the same as the sketch decoder except 1) Only one (instead of two) codebook is used and no learnable embedding is added to distinguish different codebooks in cross-attention (Eq.~\ref{eq:cross_attention}); 2) The output $h^{out}_K$ is of dimension 72 instead of 4101.

\mysubsubsection{Training} Similar to Eq.~\ref{eq:loss_fn}, reconstruction loss, codebook loss, and commitment loss are used for training. Extrude branch has a total of four code-tokens and a codebook size of $1000$.

\section{Code Selector: Details} \label{sec:appendix_code_selection}

The topology, geometry, and extrude codebooks provide $10(=4+2+4)$ codes to the sketch and extrude decoders for model generation. Without loss of generality, let us explain a topology-geometry conditioned extrude code selector. In our experiments, the maximum size of each codebook is 1,000, and we simply use a 1000 dimensional one-hot encoding to represent each code in a codebook.

Just as in the sketch decoder (Sect.~\ref{sec:decoder}), one-hot vectors are multiplied with a learnable matrix of size ($256\times 1000$) to compress feature embedding to 256 dimensional. Codes from the topology and the geometry codebooks are then injected via cross-attention to the decoder. We use the same position encoding to the codebook vectors as in Eq.~\ref{eq:cross_attention}. The decoder architecture is similar to the sketch decoder, producing the missing four extrude codes one by one in an autoregressive manner. The output embeddings finally pass through a fully-connected layer ($1000\times 256$) and softmax, predicting the probability likelihood over the $1000$ codebook indexes.

\section{Additional Results} \label{sec:appendix_experiment}

\subsection{Sketch Generation}
\label{sec:appendix_sketch}
Figure 9 to Figure 12 show the randomly generated sketches by SkexGen.

\begin{figure*}
\begin{center}
{\includegraphics[width=0.9\columnwidth]{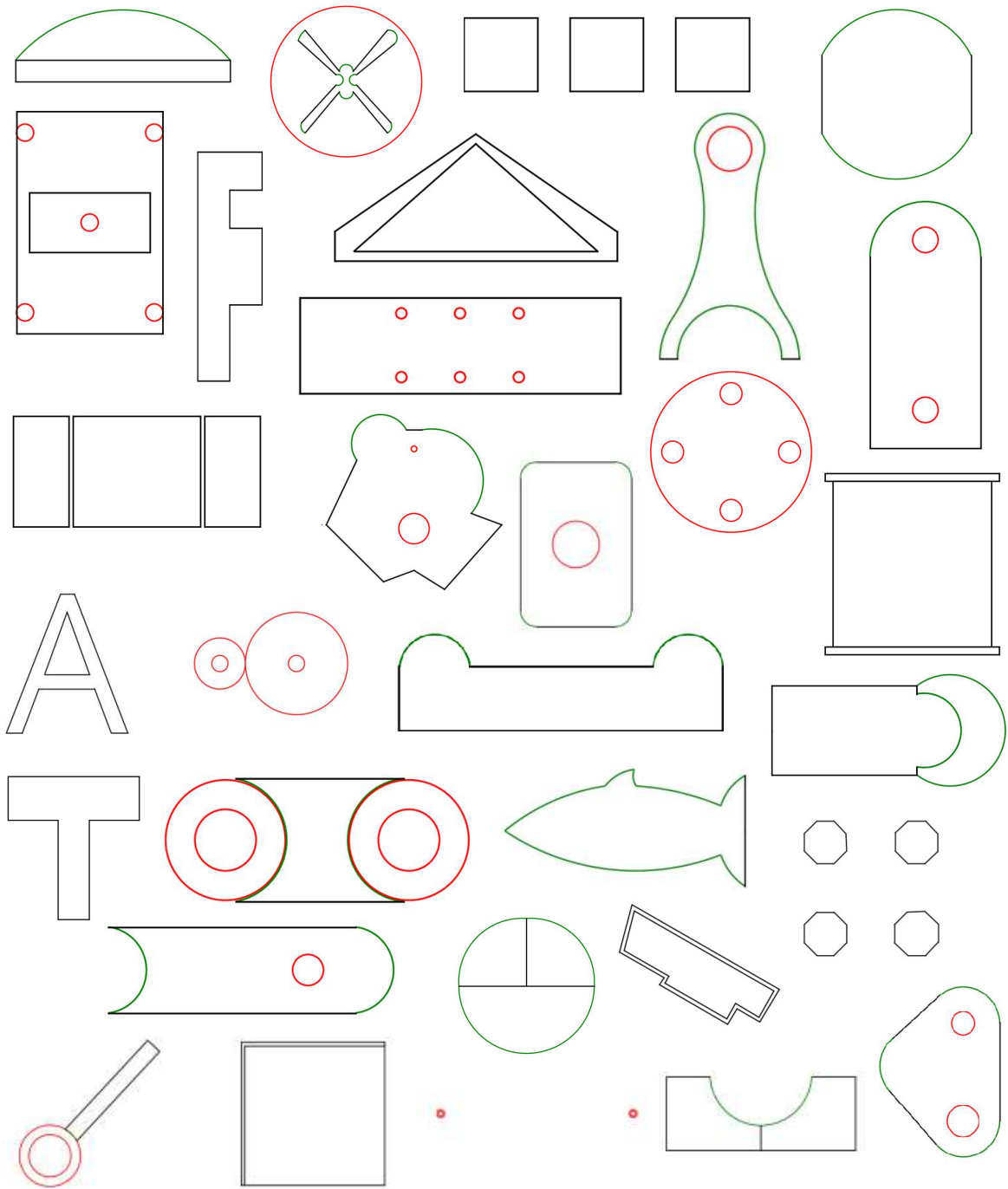}}
\caption{Randomly generated sketches by SkexGen.}
\end{center}
\end{figure*}

\begin{figure*}
\begin{center}
{\includegraphics[width=0.9\columnwidth]{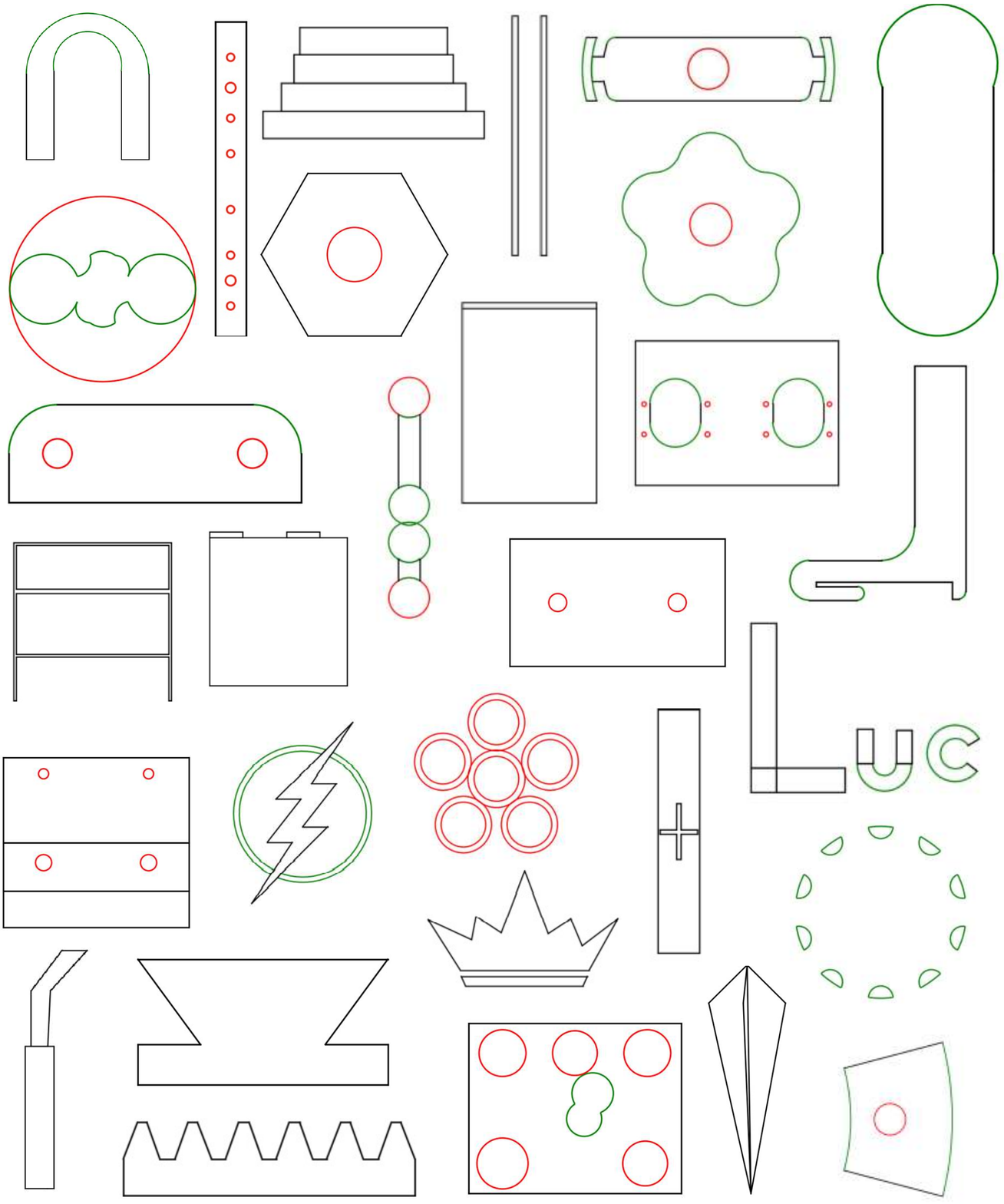}}
\caption{Randomly generated sketches by SkexGen.}
\end{center}
\end{figure*}

\begin{figure*}
\begin{center}
{\includegraphics[width=0.9\columnwidth]{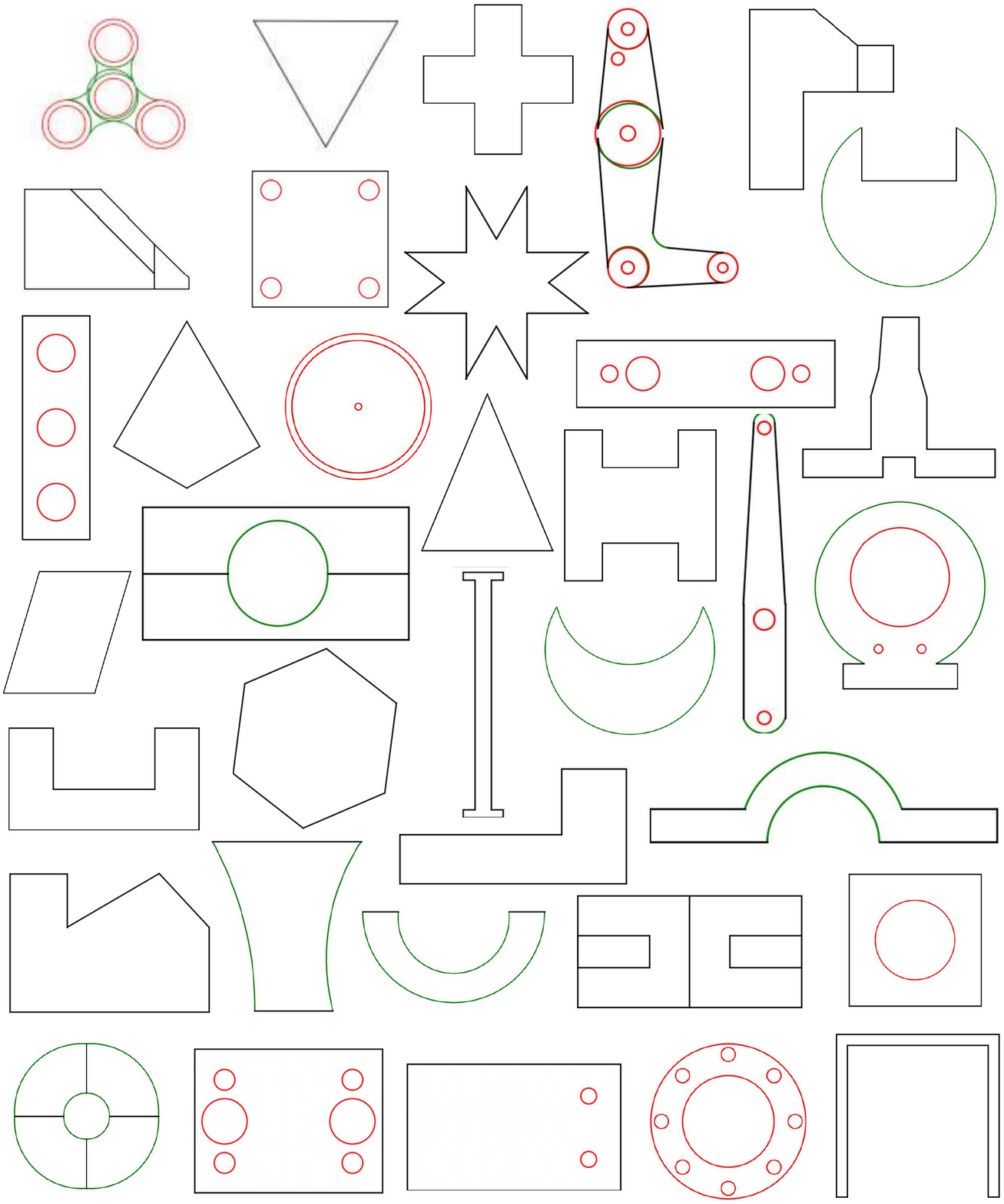}}
\caption{Randomly generated sketches by SkexGen.}
\end{center}
\end{figure*}

\begin{figure*}
\begin{center}
{\includegraphics[width=0.9\columnwidth]{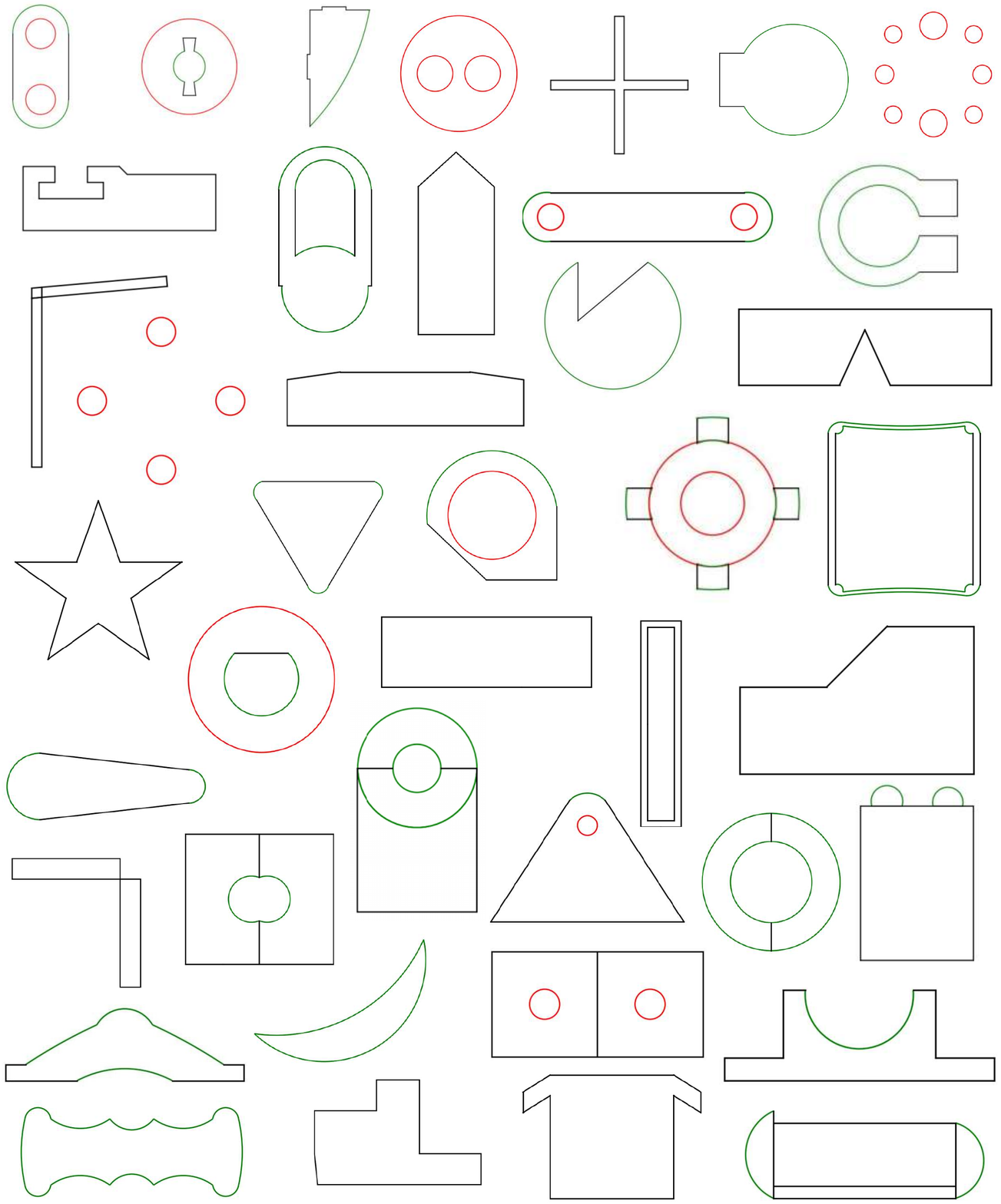}}
\caption{Randomly generated sketches by SkexGen.}
\end{center}
\end{figure*}

\subsection{CAD Generation}
\label{sec:appendix_cad}
Figure 13 to Figure 18 show the randomly generated sketch-and-extrude CAD models by SkexGen.

\begin{figure}
\begin{center}
{\includegraphics[width=0.9\columnwidth]{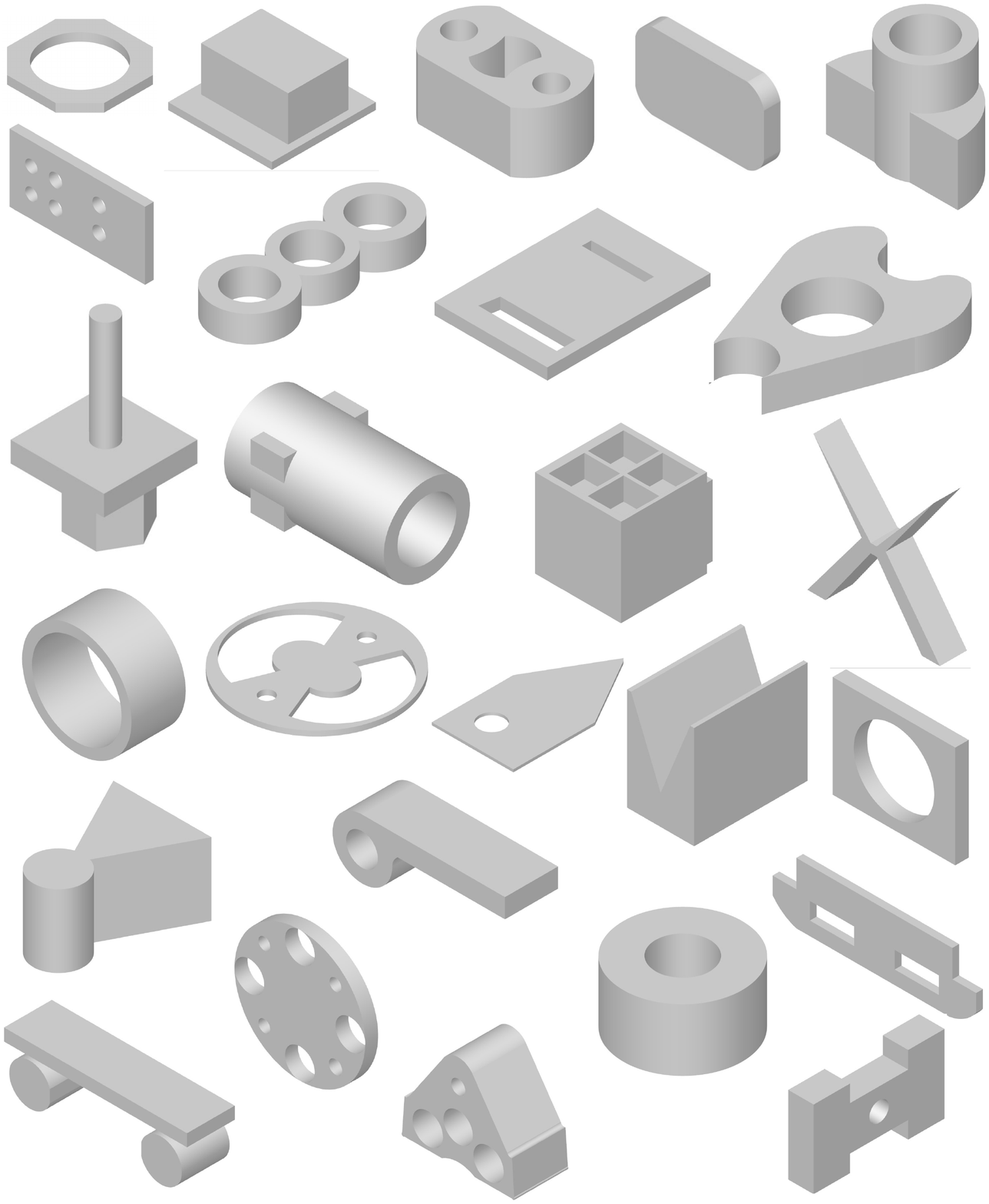}}
\caption{Randomly generated CAD models by SkexGen.}
\end{center}
\end{figure}

\begin{figure}
\begin{center}
{\includegraphics[width=0.9\columnwidth]{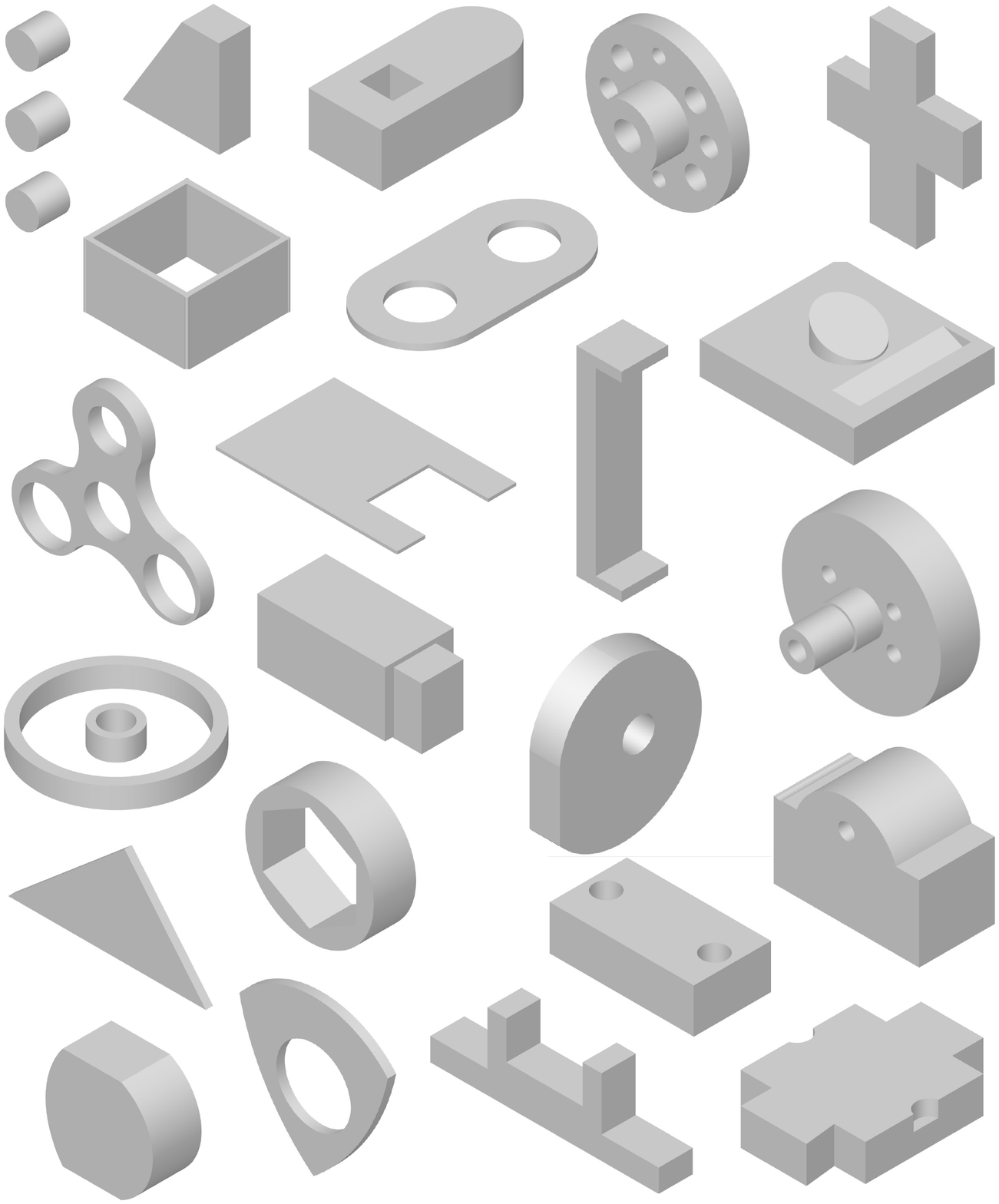}}
\caption{Randomly generated CAD models by SkexGen.}
\end{center}
\end{figure}

\begin{figure}
\begin{center}
{\includegraphics[width=0.9\columnwidth]{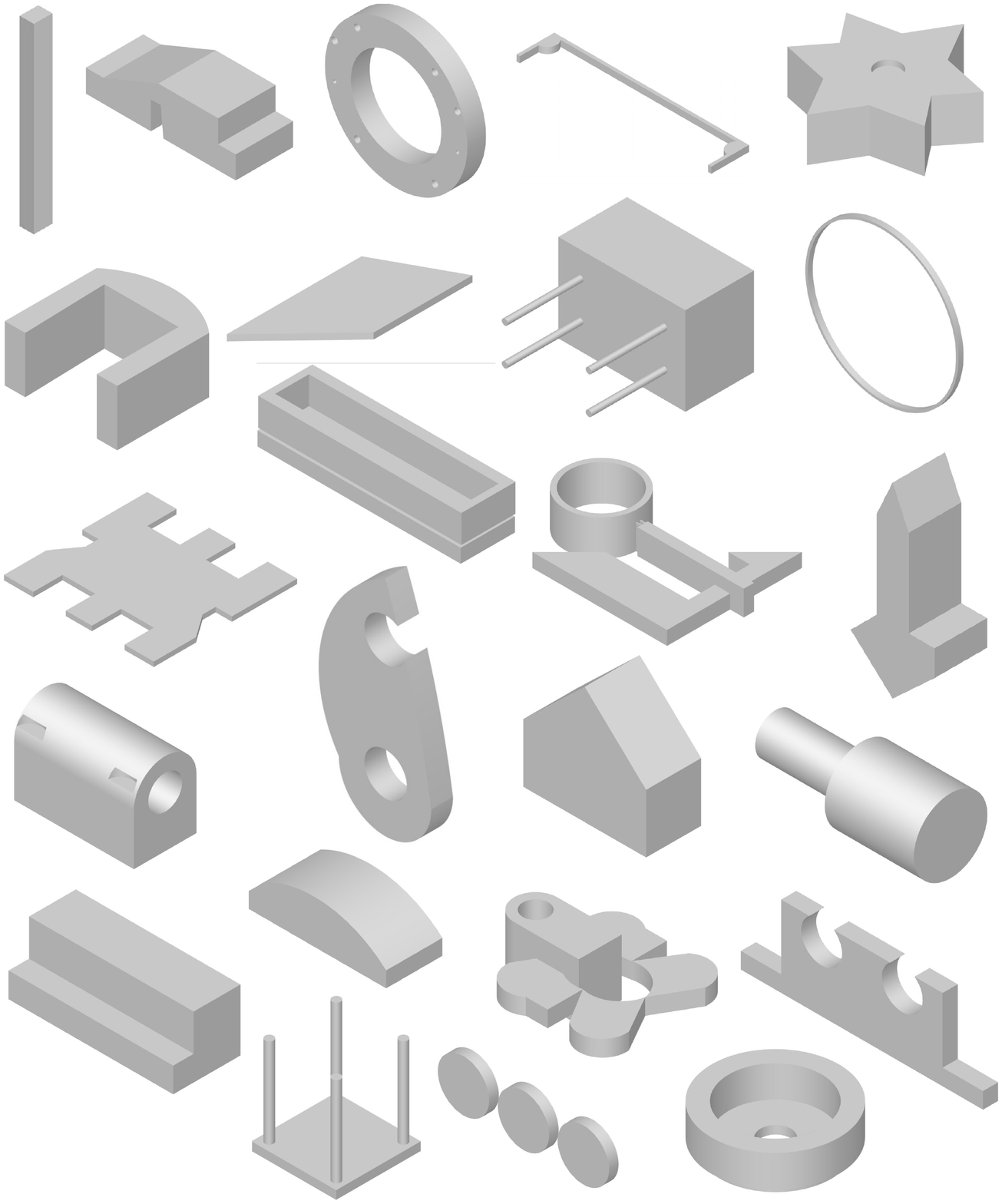}}
\caption{Randomly generated CAD models by SkexGen.}
\end{center}
\end{figure}

\begin{figure}
\begin{center}
{\includegraphics[width=0.9\columnwidth]{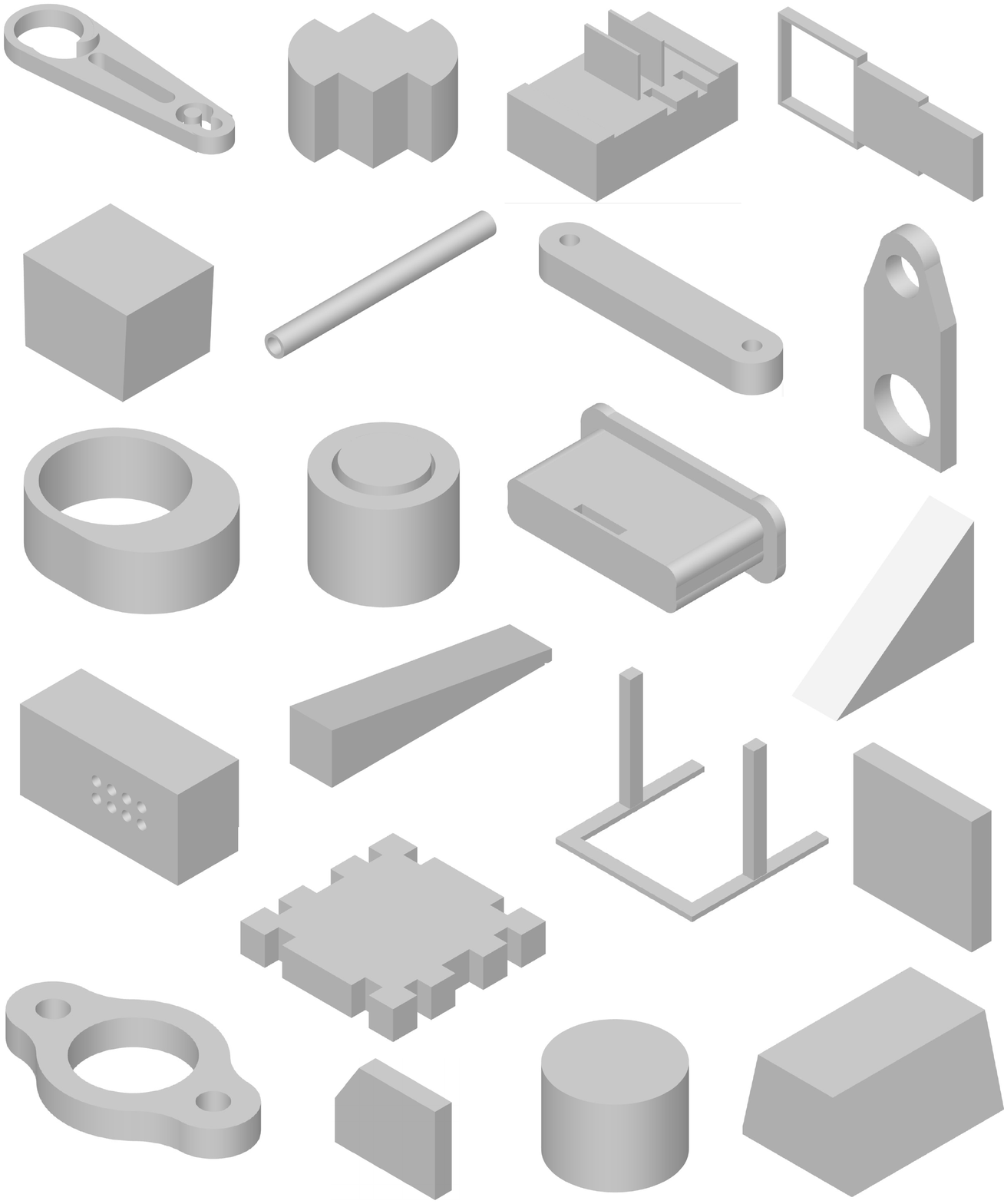}}
\caption{Randomly generated CAD models by SkexGen.}
\end{center}
\end{figure}

\begin{figure}
\begin{center}
{\includegraphics[width=0.9\columnwidth]{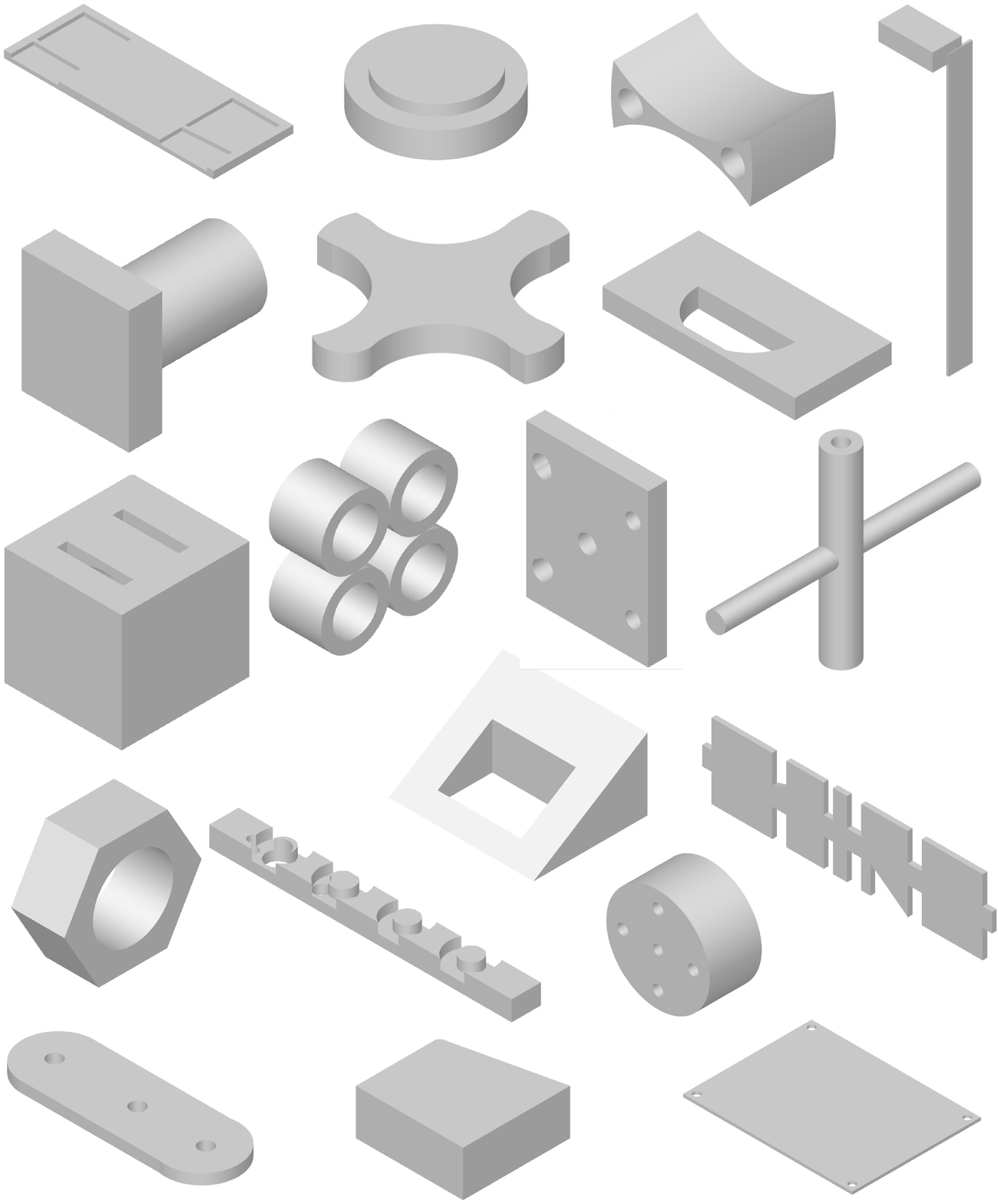}}
\caption{Randomly generated CAD models by SkexGen.}
\end{center}
\end{figure}

\begin{figure}
\begin{center}
{\includegraphics[width=0.9\columnwidth]{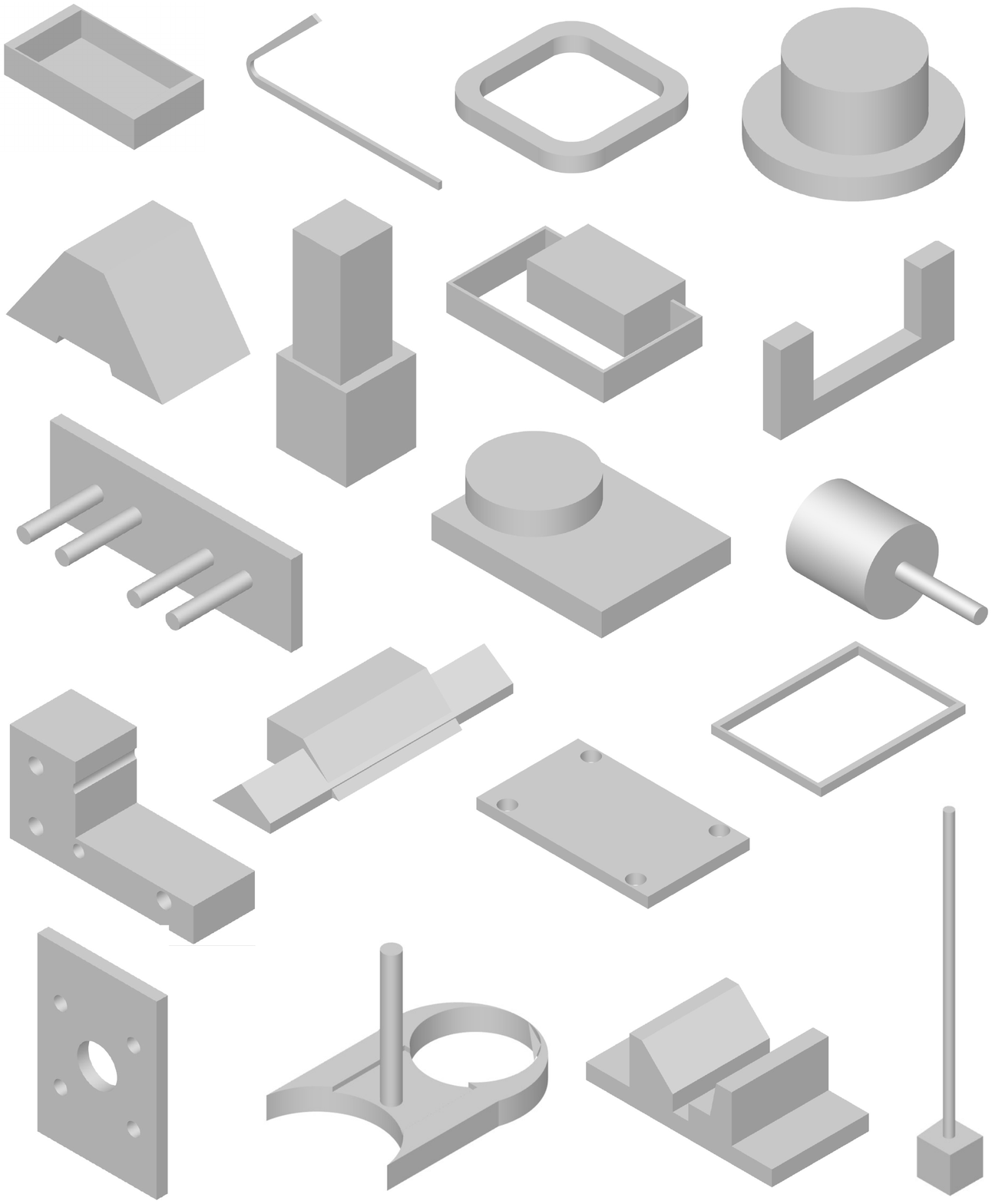}}
\caption{Randomly generated CAD models by SkexGen.}
\end{center}
\end{figure}

%%%%%%%%%%%%%%%%%%%%%%%%%%%%%%%%%%%%%%%%%%%%%%%%%%%%%%%%%%%%%%%%%%%%%%%%%%%%%%%
%%%%%%%%%%%%%%%%%%%%%%%%%%%%%%%%%%%%%%%%%%%%%%%%%%%%%%%%%%%%%%%%%%%%%%%%%%%%%%%

\end{document}